\documentclass{article}

    \usepackage[nonatbib,preprint]{neurips_2024}

\usepackage[utf8]{inputenc} 
\usepackage[T1]{fontenc}    
\usepackage{hyperref}       
\usepackage{url}            
\usepackage{booktabs}       
\usepackage{amsfonts}       
\usepackage{nicefrac}       
\usepackage{microtype}      
\usepackage{xcolor}         
\usepackage{times}
\usepackage{latexsym}
\usepackage{amssymb}
\usepackage{multirow}
\usepackage{MnSymbol}

\usepackage{subfigure}
\usepackage{graphicx}
\usepackage{float}
\usepackage{caption}
\usepackage{diagbox}
\usepackage{makecell}
\usepackage{bbding}
\usepackage{tcolorbox}
\newtcolorbox{prompt}[1]{
    left=4mm,
    right=4mm,
    top=1mm,
    bottom=1mm,
    boxsep=0mm,
    rounded corners,
    title=#1,    fontupper=\scriptsize\linespread{0.7}\fontfamily{lmr}\selectfont,
}

\usepackage{microtype}

\usepackage{inconsolata}

\usepackage{wrapfig}
\usepackage[numbers]{natbib}

\usepackage{tabularx}
\robustify\bfseries
\usepackage{subcaption}
\usepackage{cleveref}

\title{\texttt{DeepCritic}: Deliberate Critique with\\ Large Language Models}

\author{Wenkai Yang$^1$, Jingwen Chen\thanks{The work was done while Jingwen Chen was at internship in Renmin University of China.}\, $^2$, Yankai Lin\thanks{Corresponding author.} \,$^1$, Ji-Rong Wen$^1$ \\
  $^1$Gaoling School of Artificial Intelligence, Renmin University of China \\
  $^2$School of Computer Science and Technology, Beijing Jiaotong University \\
    \texttt{\{wenkaiyang, yankailin\}@ruc.edu.cn} 
    }

\begin{document}

\maketitle

\begin{abstract}
 As Large Language Models (LLMs) are rapidly evolving, providing accurate feedback and scalable oversight on their outputs becomes an urgent and critical problem. Leveraging LLMs as critique models to achieve automated supervision is a promising solution. In this work, we focus on studying and enhancing the math critique ability of LLMs. Current LLM critics provide critiques that are too shallow and superficial on each step, leading to low judgment accuracy and struggling to offer sufficient feedback for the LLM generator to correct mistakes. To tackle this issue, we propose a novel and effective two-stage framework to develop LLM critics that are capable of deliberately critiquing on each reasoning step of math solutions. In the first stage, we utilize Qwen2.5-72B-Instruct to generate 4.5K long-form critiques as seed data for supervised fine-tuning. Each seed critique consists of deliberate step-wise critiques that includes multi-perspective verifications as well as in-depth critiques of initial critiques for each reasoning step. Then, we perform reinforcement learning on the fine-tuned model with either existing human-labeled data from PRM800K or our automatically annotated data obtained via Monte Carlo sampling-based correctness estimation, to further incentivize its critique ability. Our developed critique model built on Qwen2.5-7B-Instruct not only significantly outperforms existing LLM critics (including the same-sized DeepSeek-R1-distill models and GPT-4o) on various error identification benchmarks, but also more effectively helps the LLM generator refine erroneous steps through more detailed feedback.\footnote{Data and models are available at \url{https://github.com/RUCBM/DeepCritic}.}
\end{abstract}

\section{Introduction}

Large Language Models~(LLMs)~\citep{gpt4, qwen2.5, deepseek-v3} have demonstrated superior performance that even surpasses human capabilities across a wide range of tasks~\citep{swebench,llm_predict_neuroscience,o1}. LLMs achieve strong and generalizable performance by training on human-provided knowledge data~\citep{instructGPT}. This makes their evolution highly dependent on the effective human supervision~\citep{hh-rlhf}. However, as LLMs become increasingly intelligent, providing effective and scalable human supervision will become highly challenging, as collecting human feedback would be too costly and difficult~\citep{scalable-oversight}. 

LLM critics~\citep{llm-critics-help-catch-llm-bugs}, which leverage LLMs as critique models, have recently emerged as a promising approach to enabling scalable oversight~\citep{scalable-oversight} on evolving LLMs. LLM critics generate critiques of LLM-generated content, which identify flaws and errors to help the LLM generator refine its outputs, paving the way for the automatic supervision and continuously improvement of LLMs~\citep{self-refine,critiquellm,llm-critic-catch-math-bugs,scrit}.

\begin{figure*}[t!] 
    \centering
    \includegraphics[width=\textwidth]{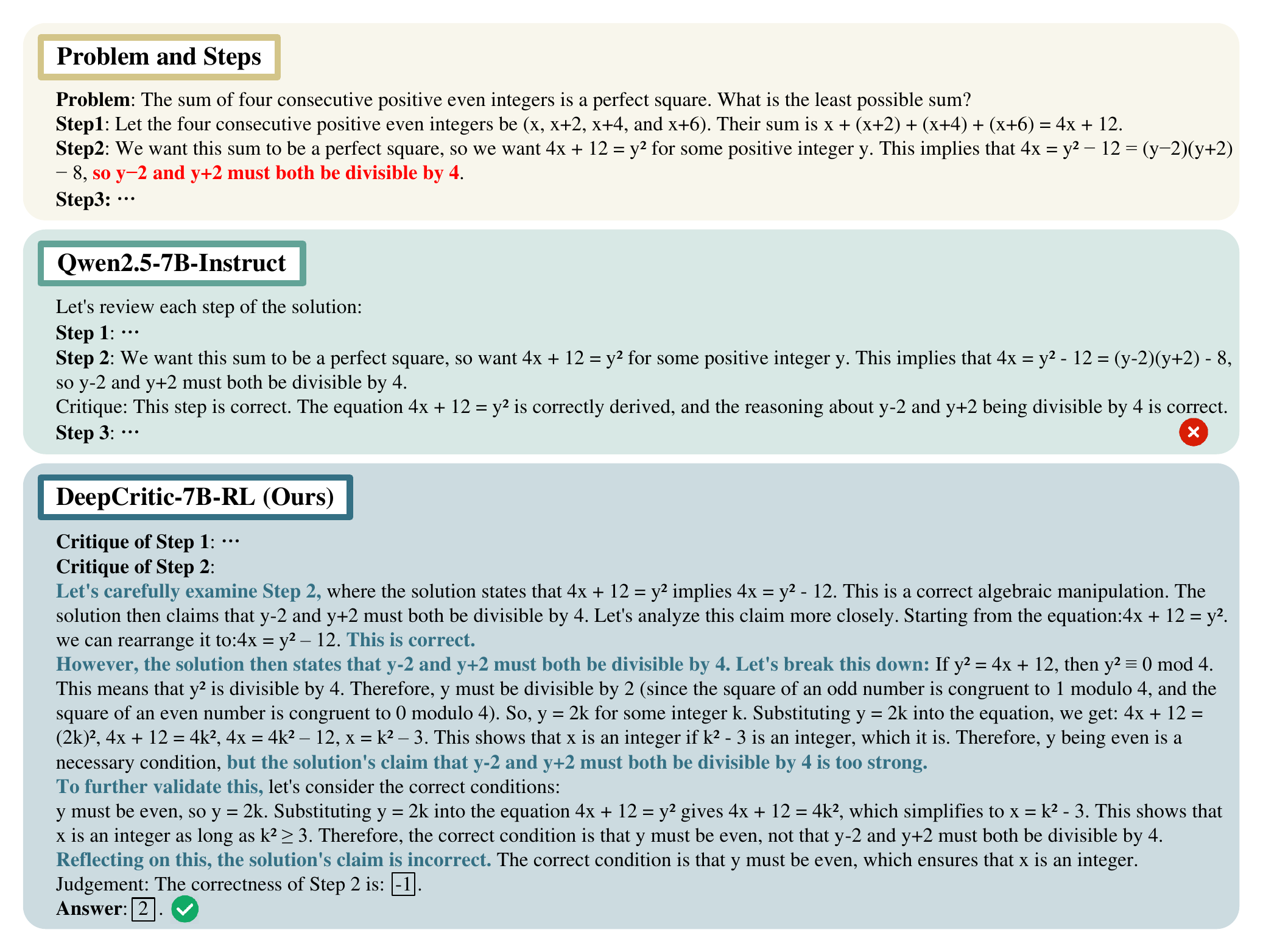}
    \caption{Comparison of critiques generated by current LLM critics and our developed critic. The red highlights in the solution steps represent the erroneous part. The critiques of current LLM critics (e.g., Qwen2.5-7B-Instruct) are overly superficial, primarily consisting of declarative statements rather than in-depth analysis or critical evaluation. In contrast, our critic can generate a deliberate reasoning process before making a judgment, incorporating \textbf{iterative evaluation}, \textbf{multi-perspective verification}, and \textbf{meta-critiquing}.
    }
    \label{fig: comparison}
\end{figure*}

The effectiveness and the potential of the utilizing LLM critics heavily depends on the inherent critique capabilities of the LLM critics. However, existing studies have shown that current LLM critics still exhibit limited critique capabilities in complex domains such as mathematical reasoning tasks~\citep{Mr-gsm8k,Processbench}, making it difficult to provide accurate and reliable feedback. By analyzing the critiques generated by existing LLM critics on math problems and solutions, we find that their generated critiques are often overly superficial and lack critical thinking, as shown in the examples in Figure~\ref{fig: comparison}. In particular, their critiques merely follow and echo the same reasoning process of the original steps, re-iterating rather than challenging them with their own critical reasoning, leading to a premature judgment result. This behavior can lead to two problems: (1) The critiques of reasoning steps lack careful deliberation, leading to low accuracy in the judgment results. (2) The critiques lack informativeness, offering limited guidance for the LLM generator to perform targeted refinements.

In this work, we aim to address the aforementioned limitations of shallow critiques generated by the LLM critics, particularly in the domain of mathematical reasoning. 
Specifically, we propose the \textbf{\texttt{DeepCritic}} framework, which enhances the critique capabilities of LLMs through a two-stage training pipeline. 
In the first stage, to enable LLMs to acquire a preliminary capability for generating fine-grained critiques, we first curate a dataset of 4.5K long-form critiques by iteratively prompting Qwen2.5-72B-Instruct~\citep{qwen2.5} to critique on a small subset of labeled solutions in PRM800K~\citep{verify_step_by_step}. Each above critique includes step-wise critiques of all reasoning steps if the solution is correct, or up to the first erroneous step otherwise. 
When constructing each step-wise critique, we first generate an initial and preliminary critique of the specified reasoning step. Then, in order to enable our critic to conduct critiques more critically and from more diverse perspectives, we further generate an in-depth critique of the initial critique. The in-depth critique is supposed to validate the step from alternative perspectives and critically examine the initial critique itself. Finally, we merge the initial and in-depth critique into one deliberate critique for the specified step. By supervised fine-tuning on the curated and filtered critique data, we obtain an initial critique model that is capable of performing multi-perspective evaluation and meta-critiquing through reflection on and correction of its prior erroneous judgments. Then, in the second stage, we perform reinforcement learning~(RL) to the SFT model to further boost its deep critique ability. We perform RL under two different settings based on the source of RL data: (1) When human-labeled data is available, such as PRM800K, we directly use it for RL; (2) Otherwise, we automatically generate annotated RL data through Monte Carlo sampling-based correctness estimation on each reasoning step~\citep{Math-shepherd} and then perform RL.

Experimental results show that our developed deep critic significantly outperforms existing process reward models~(PRMs) and LLM critics (including the advanced R1-distilled reasoning models~\citep{r1} and GPT-4o~\citep{gpt4}) on various error identification benchmarks, demonstrating the effectiveness of our pipeline in enabling LLM critics to provide more accurate supervision. Furthermore, we demonstrate promising test-time scaling properties for both our deep critic and the LLM generator within our framework: (1) the judgment accuracy of the critic consistently improves with increased test-time sampling effort; and (2) the performance of the LLM generator is effectively enhanced either by employing our deep critic as a verifier in majority voting or through refinement guided by the critic's more informative feedback.


\section{Related Work}
\textbf{Critique Ability of LLMs}
In the current era of rapidly evolving LLMs, leveraging and improving the critique ability of LLMs to facilitate effective scalable oversight~\citep{measureing_scalable_oversight,scalable-oversight} and superalignment~\citep{weak-to-strong, deceive_weak_models} has become increasingly important. Regarding the utilization of the critique ability of LLMs, LLM-as-a-Judge~\citep{survey-on-llm-as-a-judge} and LLM-as-a-Critic~\citep{llm-critics-help-catch-llm-bugs} serve as promising directions for facilitating automatic supervision of LLM generations~\citep{llm-as-a-judge}, enabling the self-evolution of LLMs~\citep{self-rewarding,meta-rewarding}, and refining LLM outputs at test time~\citep{self-refine}. Benchmarking the critique ability of LLMs~\citep{Critique_ability,criticeval,criticbench} paves the way to better understanding the potential and limitations of current LLMs on critique tasks. Finally, a series of studies aim to create more powerful critique models on mathematical reasoning~\citep{generative_prm,critic-cot, llm-critic-catch-math-bugs, enhancing-llm-via-critique, scrit}, code generation~\citep{llm-critics-help-catch-llm-bugs,ctrl} or other open-domain tasks~\citep{critiquellm}. This work analyzes the limitations of current math critique models and proposes a novel and effective pipeline to enhance the math critique ability of LLMs.~\looseness=-1

\noindent \textbf{Reasoning Ability of LLMs}
The reasoning abilities of LLMs has long been a topic of widespread interest in the community. Previous studies explore LLM reasoning in various domains, such as code~\citep{codellama,qwen2.5-coder}, math~\cite{deepseek-v3,qwen2.5-math}, commonsense knowledge~\citep{gpqa, strategyqa}, etc. This work mainly focuses on the math reasoning domain. The studies in this line can be divided into several categories: (1) Designing diverse and challenging math reasoning benchmarks to probe the boundaries of existing LLMs' reasoning abilities~\citep{gsm8k, math, omni-math}. (2) Enhancing the math reasoning abilities of LLMs in the training time either by collecting high-quality math datasets for fine-tuning~\citep{metamath,numinamath}, or by proposing advanced optimization algorithms~\citep{step-dpo,prime}. (3) Improving reasoning accuracy by increasing the test-time compute either by performing search-based sampling~\citep{self-consistency,compute-optimal-scaling} with process reward models~(PRMs)~\citep{verify_step_by_step,Math-shepherd}, or by extending the Chain of Thought~(CoT) length~\citep{o1,r1,Thinking-Optimal_Scaling}. 
This work, taking a pioneer step to have the critique model provide judgments after detailed and deliberate reasoning, can also improve LLMs' math reasoning by providing accurate and detailed feedback on erroneous solutions and assisting LLMs in correcting them.

\section{Methodology}

\subsection{Problem Formulation}
Here, we introduce the preliminaries of our studied critique problem setting. Let $\mathcal{D}=\{ (P,S) \}$ denote a dataset comprising pairs of problems $P$ and their corresponding solutions $S$. Each solution $S$ is in a step-by-step format denoted as $S=(s_{1}, s_{2},\cdots, s_{n})$, where $s_{i}$ represents the $i$-th step. Denote $\boldsymbol{\theta}_{c}$ as an LLM critic whose role is to review each step in $S$ sequentially, identify in which step the first error occurs, and return the step index of that first erroneous step. If all steps are deemed correct, it returns -1 to indicate that the entire solution is correct. Formally, the output of the critic can be formulated as
\begin{equation}
\label{eq: critic output}
\begin{aligned}
& (c_{1},j_{1},\cdots, c_{k},j_{k}, a)\sim \pi_{\boldsymbol{\theta}_{c}}(\cdot | P, s_{1}, \cdots, s_{n}),
\end{aligned}
\end{equation} 
where $c_{i}$ represents the CoT critique of step $s_i$, $j_{i} \in \{1,-1 \}$ represents the judgment result (i.e., 1 for correct or -1 for incorrect) indicating the correctness of $s_i$. The final judgment result $a$ is equal to $k$ if $j_k = -1$, indicating the first incorrect step; otherwise, if all $j_i = 1$ for $i \leq k = n$, then $a = -1$.

As mentioned before (refer to Figure~\ref{fig: comparison}), current LLM-based critique models exhibit limitations in conducting thoughtful critiques, as their step-wise critiques $c_i$ tend to be overly brief and superficial, often echoing the original reasoning content without deeper and critical analysis. Thus, their critique performance is greatly limited as shown in previous studies~\citep{Mr-gsm8k,Processbench}, and the shallow, uninformative critiques fail to provide useful guidance for the LLM generator to revise its solutions. In this work, we aim to improve the LLM's critique ability and enable it to produce more deliberate and thoughtful critiques $c_{i}$ before making the judgment result $j_{i}$, enhancing both the accuracy and quality of its generated critiques.

\subsection{\texttt{DeepCritic}: Deliberate Critique Model}
In this section, we will introduce our two-stage pipeline to create deliberate critique models in detail, including the SFT data generation and training stage in Section~\ref{subsubsec: sft}, and the RL data curation and optimization stage in Section~\ref{subsubsec: rl}.

\begin{figure*}[t!] 
    \centering
    \includegraphics[width=\textwidth]{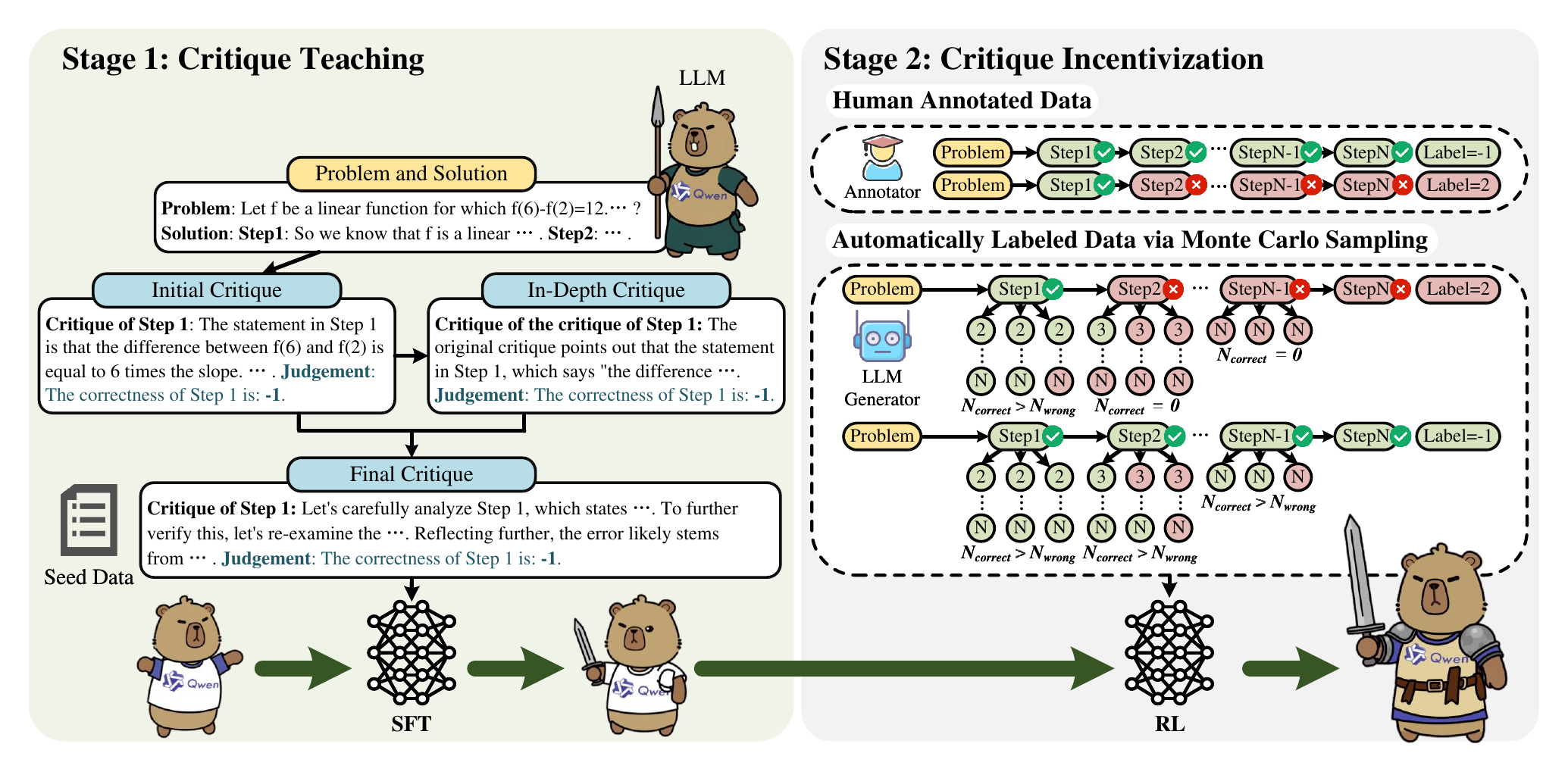}
    \caption{The two-stage pipeline of training our deep critique models. In Stage 1, we first utilize Qwen2.5-72B-Instruct to generate an initial step-wise critique for each step in the solution, followed by an in-depth critique of the initial critique. Then, we use Qwen2.5-72B-Instruct to merge these two critiques into one deliberate critique in the long-CoT form. Finally, we perform SFT on above created critique data to teach the model the format of deliberately critiquing. In Stage 2, we perform RL to the SFT model on either existing human-annotated data or auto-labeled data via Monte Carlo sampling-based correctness estimation, to further stimulate the critique ability of the critic.
    }
    \label{fig: pipeline}
\end{figure*}

\subsubsection{Teaching LLMs to Deliberately Critique}
\label{subsubsec: sft}
Given that existing LLM critics struggle to produce well-reasoned and in-depth critiques, our first stage aims to \textbf{\textit{teach}} LLMs how to deliberately critique. In this stage, we first leverage Qwen2.5-72B-Instruct~\citep{qwen2.5} to iteratively perform initial and in-depth critiquing, and then merge the two critiques into a single long-form critique to serve as the seed critique. We subsequently perform SFT on the curated seed critique data to teach the target model the format and structure of deliberate critiquing. The brief illustration is displayed in the left part of Figure~\ref{fig: pipeline}, and the detailed procedure is described below.

\textbf{Initial Critique Generation} First, we sample a small set of labeled data from PRM800K~\cite{verify_step_by_step} as the seed task inputs. Each task input contains a problem $P$ and a step-by-step solution $S=(s_1,\cdots,s_n)$, along with the human-labeled label $l_{i}\in \{ 1, -1\}$\footnote{In the original PRM800K dataset, there are some steps labeled with 0, indicating that these steps is not incorrect but do not make any progress. We consider the label for these steps to be 1.} indicating the correctness of each reasoning step $s_i$. Thus, 
we can create the step-by-step critiques on these seed inputs using an LLM $\boldsymbol{\theta}^{*}$, which is chosen as Qwen2.5-72B-Instruct in this work. However, instead of creating the step-by-step critique of the entire solution directly in a single pass just like Eq.~(\ref{eq: critic output}), which often leads current LLMs to generate overly brief critiques for each step as mentioned before, we adopt an approach that critiques each step independently. Specifically, in each prompting of Qwen2.5-72B-Instruct, we provide the problem and entire solution as inputs, but instruct Qwen2.5-72B-Instruct to critique only one specified step:
\begin{equation}
\label{eq: step-level critique}
\begin{aligned}
(c_{i}^{init}, j_{i}^{init}) = \pi_{\boldsymbol{\theta}^{*}}(\cdot | P, s_1, \cdots, s_{n}, s_{target}=s_i), \quad i=1,\cdots, k,
\end{aligned}
\end{equation} 
where $s_{target}$ is the additional requirement that specifies the $i$-th step to be critiqued only, $(c_{i}^{init}, j_{i}^{init})$ represents the initial CoT critique and judgment result of the specified step, $k$ represents the index of the first step where $l_i = -1$, or $k=n$ when all $l_{i}$ is 1. 

\textbf{In-Depth Critique Generation} After initial critique generation, we find that many of the initial critiques merely adopt a direct verification approach that directly follows the logic of the original reasoning steps and perform repetitive or similar calculations, resulting in relatively low accuracy when identifying incorrect steps. To enable our critique model to learn to perform critical evaluations, we introduce a second round of in-depth critique generation based on the initial critiques. Specifically, for each reasoning step in the solution, we instruct Qwen2.5-72B-Instruct again to either \textbf{assess the reasoning step from a different perspective or using a different evaluation method} than that used in the initial critique, or to \textbf{critique the initial critique itself} in order to identify whether there exist flaws in the initial critique that lead to the incorrect judgment about the reasoning step. Therefore, the in-depth critique $c^{deep}_{i}$ and its judgment result $j^{deep}_{i}$ are generated as
\begin{equation}
\label{eq: critique of critique}
\begin{aligned}
(c_{i}^{deep}, j_{i}^{deep}) = \pi_{\boldsymbol{\theta}^{*}}(\cdot | P, s_1, \cdots, s_{n}, c_{i}^{init}, j_{i}^{init}, s_{target}=s_i), \quad  i=1,\cdots, k.
\end{aligned}
\end{equation} 
This process allows initial critiques that previously led to mismatches between the initial judgment result and the ground truth label (i.e., $j^{init}_{i} \neq l_{i}$) to be revised into correct critiques. Then, we only retain the solutions in which the in-depth judgment results of all steps in the solution align with the ground truth labels (i.e., $j^{deep}_{i} = l_{i}, \forall i=1,\cdots,n$), as well as their initial and in-depth critiques. Table~\ref{tab: proportion of corrected steps} shows the proportion of step-level critiques in our final SFT dataset that are successfully corrected through the second-round in-depth critique generation process. We observe that a certain number of step-level critiques benefit from the in-depth critique generation process. Incorporating these samples into training equips the model with the capabilities of reflection and self-correction in critiquing.

\begin{table*}[t]
\caption{Statistics of step-level critiques in our SFT dataset, categorized by the correctness of their corresponding initial critiques.}
\label{tab: proportion of corrected steps}
\centering
\begin{tabular}{ccc}
\toprule
Label of Reasoning Step & \# Correct Initial Critiques & \# Incorrect Initial Critiques \\
\midrule
1 & 22968 & 738 \\
-1 & 3535 & 565 \\
\bottomrule
\end{tabular}
\end{table*}

\textbf{Final Critique Synthesis} In the last step of SFT data generation, we use in-context learning with two manually-written examples to instruct Qwen2.5-72B-Instruct to merge the initial and in-depth critiques of each step into a single deliberate critique:
\begin{equation}
\label{eq: critique merge}
\begin{aligned}
(c_{i}^{final}, j_{i}^{final}) = \pi_{\boldsymbol{\theta}^{*}}(\cdot | c_{i}^{init}, j_{i}^{init},c_{i}^{deep},j_{i}^{deep}, \{ex_{l}\}), \quad  i=1,\cdots, k,
\end{aligned}
\end{equation} 
where $\{ex_{l} \}$ are in-context learning examples. 
Finally, we only need to concatenate all step-level critiques to form a complete solution-level critique: 
\begin{equation}
\label{eq: final critique}
\begin{aligned}
C = (c^{final}_{1}, j^{final}_{1}, \cdots, c_{k}^{final}, j^{final}_{k}, a), \quad a = \begin{cases}
k & \text{if } j^{final}_{k}=-1, \\
-1 & \text{if } j^{final}_{k}=1.
\end{cases}
\end{aligned}
\end{equation} 
Such deliberate critiques enable the model to \textbf{perform iterative evaluations, multi-perspective verifications, reflection, and meta-critiquing} in the inference stage, thereby improving its judgment accuracy.  
All prompt templates used in the seed data generation stage and the corresponding generation hyper-parameters can be found in Appendix~\ref{appendix: prompt templates} and Appendix~\ref{appendix: hyper-parameters in critique generation} respectively.

\textbf{Supervised Fine-Tuning} In total, we obtain approximately 4.5K seed solution-level critiques, and their label distribution (i.e., the distribution of the step index of the first erroneous step) is shown in Figure~\ref{fig: label distribution in sft data}. We then perform SFT on the target model  to teach it the format for performing deliberate critique and obtain an initial critique model $\boldsymbol{\theta}_{SFT}$:
\begin{equation}
\label{eq: sft}
\begin{aligned}
\boldsymbol{\theta}_{SFT} = \mathop{\arg\min}_{\boldsymbol{\theta}} \mathbb{E}_{(P,S,C)\sim \mathcal{D}_{SFT}}[- \log P_{\boldsymbol{\theta}}(C|P,S)],
\end{aligned}
\end{equation} 
where $\mathcal{D}_{SFT}$ is the SFT critique dataset in which the input includes the problem and the solution, and the output is the solution-level deliberate critique.

\subsubsection{Incentivizing LLMs to Deliberately Critique}
\label{subsubsec: rl}
Once the seed critique model has acquired a certain level of critique capability, in the second stage, we aim to stimulate and elicit its full potential through continued \textbf{\textit{incentivization}}. We follow the recent exciting advancements in LLM reasoning domain~\citep{o1,r1,open-reasoner-zero} to employ reinforcement learning~(RL) on $\boldsymbol{\theta}_{SFT}$ in the second stage's training.

The acquisition of RL data is critical in RL stage. In the following, we explore RL under two different settings based on the sources of data. (1) First, the ideal source for RL should ne the high-quality labeled data obtained through human annotation. Therefore, in the first setting we directly use the existing human-labeled dataset PRM800K~\citep{verify_step_by_step} for RL. (2) However, in some cases human annotation may become impractical or even infeasible due to high cost. Thus, in the second setting where human annotation is infeasible, we construct the task data automatically via a Monte Carlo sampling-based correctness estimation method~\citep{Math-shepherd}. Specifically, we sample a portion of GSM8K, MATH and Olympiads problems from NuminaMath-CoT dataset~\citep{numinamath}, and leverage Qwen2.5-1.5B/3B/7B-Instruct~\citep{qwen2.5} to generate multiple step-by-step solutions for each problem. Problems where all solutions are either fully correct or fully incorrect are discarded, as such cases are deemed too easy or too challenging. Then, for each incorrect solution, to measure the correctness of a specific step $s_{i}$, we follow~\citet{Math-shepherd} to truncate the solution after $s_i$, and use an LLM generator (i.e., Qwen2.5-7B-Instruct in this work) to rollout the subsequent reasoning path $N$ times independently. \textbf{We define the first erroneous step as the first step from which, along with all subsequent steps, all rollouts generated by the generator are incorrect; while for its all preceding steps, more than half of rollouts reach the correct answers.} If such steps do not exist, we discard those incorrect solutions. For solutions with correct final answers, prior studies~\citep{Processbench, scrit} have pointed out that their intermediate steps can still be incorrect. Therefore, we perform the same Monte Carlo sampling procedure and \textbf{assign a label of -1 only to those correct solutions whose all intermediate steps have corresponding rollouts where more than half reach the correct answers}.

The illustration of above data construction procedure is shown in the right part of Figure~\ref{fig: pipeline}. The detailed data generation settings are put in Appendix~\ref{appendix: hyper-parameters in critique generation}. In our experiments, we explore training the seed critique model with RL either using 40.7K PRM800K data or 14.2K automatically constructed data. The label distributions of these two data sources are shown in Figure~\ref{fig: label distribution in prm800k data} and Figure~\ref{fig: label distribution in numina data} respectively.

\section{Experiments and Analysis}
\subsection{Experimental Settings}

\textbf{Base Model} We choose Qwen2.5-7B-Instruct as the initial base model. We first perform SFT to get our seed critique model \textbf{DeepCritic-7B-SFT}. Then, we perform RL on two distinct types of RL data separately, resulting in two variants: \textbf{DeepCritic-7B-RL-PRM800K} and \textbf{DeepCritic-7B-RL-Numina}.

\textbf{Benchmarks} We select three widely used error identification benchmarks to systematically evaluate the critique and judgment performance of each model, including the subset of \textbf{MR-GSM8K}~\citep{Mr-gsm8k} in which the questions are from original GSM8K~\citep{gsm8k} dataset, the Phase-2 test set\footnote{\url{https://github.com/openai/prm800k/blob/main/prm800k/data/phase2_test.jsonl}} of \textbf{PRM800K}~\citep{verify_step_by_step}, and \textbf{ProcessBench}~\citep{Processbench}. Each testing example in all datasets contains a problem, a step-by-step solution and a label that either represents the step index of the first erroneous step or is -1 if the solution is entirely correct. The detailed description of the three benchmarks is provided in Appendix~\ref{appendix: introduction to benchmarks}.

\textbf{Baselines} We compare our critique models against two categories of baselines: (1) \textbf{Process Reward Models}~(PRMs): In this category, we select Math-Shepherd-PRM-7B~\citep{Math-shepherd}, RLHFlow-PRM-8B-Mistral/DeepSeek~\citep{rlhflowmath}, Qwen2.5-Math-7B-PRM800K~\citep{Processbench} for comparison. (2) \textbf{LLM Critics}: We prompt the following leading LLMs to serve as critique models: LLaMA3.1-8B/70B-Instruct~\citep{llama3.1}, Qwen2.5-7B/72B-Instruct~\citep{qwen2.5}, Qwen2.5-Math-7B/72B-Instruct~\citep{qwen2.5-math}, and GPT-4o~\citep{gpt4o}. Also, we include two of the most advanced reasoning LLMs, DeepSeek-R1-Distill-Llama-8B and DeepSeek-R1-Distill-Qwen-7B~\citep{r1}, and use them as reasoning-enhanced critique models for comprehensive comparison.

\textbf{Training Settings} In SFT stage, the learning rate is $1\times 10^{-5}$, the batch size is $64$, and we fine-tune for 3 epochs. In RL stage, we adopt \texttt{verl}~\citep{verl} as our training framework, and use Group Relative Policy Optimization~(GRPO)~\citep{deepseekmath} as RL algorithm.  In RL, an accuracy reward of 1.0 is given if the final judgment is correct; otherwise, it is
0.0. During RL, we observe that in very few cases, the policy model generates critiques that are mixed with different languages, which is consistent with the findings in DeepSeek-R1~\citep{r1}. However, this issue gradually diminishes as training progresses, so we do not introduce a language consistency reward here. The detailed training settings in both SFT and RL stages are in Appendix~\ref{appendix: training setting}.

\textbf{Evaluation Settings} In our main evaluation, we use consistent sampling settings across all critique models, with \texttt{temperature} set to 0.6, \texttt{top\_p} to 0.9, and \texttt{max\_generation\_length} to 32K during inference. We only sample once for each task input, while we explore the performance of majority voting over eight samplings in Section~\ref{subsec: majority voting}. The evaluation prompt is mainly based on~\citep{Processbench}, and is put in Appendix~\ref{appendix: evaluation settings}. The main evaluation metric is the F1 score~\citep{Processbench}, which is the harmonic mean of the judgment accuracy on the step index of first erroneous step in incorrect solutions and the judgment accuracy on correct solutions. Further details on the evaluation settings are provided in Appendix~\ref{appendix: evaluation settings}.

\subsection{Main Results}
\begin{table*}[t]
\caption{The evaluation results of various PRMs, instruction-followed LLMs that are served as critique models and our critique models on three benchmarks assessing the mathematical critique ability. The reported metric is the F1 score~\citep{Processbench} (i.e., harmonic mean) of the judgment accuracy on incorrect solutions and the judgment accuracy on correct solutions.}
\label{tab: main results}
\centering
\small
\setlength{\tabcolsep}{3.25pt}
\begin{tabular}{lccccccc}
\toprule
\multirow{3.5}{*}{\begin{tabular}[c]{@{}l@{}}Model \end{tabular}} &  \multirow{3.5}{*}{\begin{tabular}[c]{@{}c@{}}MR-\\GSM8K\end{tabular}} & \multirow{3.5}{*}{PRM800K} & \multicolumn{4}{c}{ProcessBench}  & \multirow{2.5}{*}{Avg.} \\
\cmidrule(lr){4-7}
&   &   &  \multirow{2}{*}{GSM8K}  &  \multirow{2}{*}{MATH}    & \multirow{2}{*}{\begin{tabular}[c]{@{}c@{}}Olympiad-\\Bench\end{tabular}} &  \multirow{2}{*}{Omni-Math} &   \\
& \\
\midrule
\multicolumn{8}{l}{\emph{\quad \textbf{Process Reward Models} (PRMs)}} \\
Math-Shepherd-PRM-7B & 61.8 & 21.7 & 48.2 & 27.1 & 20.5 & 16.3 & 32.6 \\
RLHFlow-PRM-8B-Mistral & 66.6 & 25.2 & 50.9 & 32.0 &  13.8 & 15.7 & 34.0 \\
RLHFlow-PRM-8B-DeepSeek &  44.8 & 18.5 & 32.3 & 34.2 & 16.0 & 18.3 & 27.4\\
Qwen2.5-Math-7B-PRM800K & 70.8 & 55.6 & 70.5 & 64.7 & 50.0 & 42.7 & 59.7\\
\midrule
\multicolumn{8}{l}{\emph{\quad Large Language Models, served as \textbf{Critique Models}}} \\
LLaMA3.1-8B-Instruct & 31.6 & 16.0 &  23.8 & 18.9 &  18.3 & 17.2 & 21.0 \\
Qwen2.5-7B-Instruct & 48.1 & 25.6 &  42.9 & 36.6 &  25.5 & 25.9 & 34.1 \\
Qwen2.5-Math-7B-Instruct & 35.6 & 19.4 &  23.1 & 22.0 & \phantom{0}9.2 & 10.4 & 20.0 \\
DeepSeek-R1-Distill-Llama-8B & 69.4 & 55.7 & 65.0 & 62.7 & 58.4 & 51.7 & 60.5 \\
DeepSeek-R1-Distill-Qwen-7B & \textbf{77.9} & 57.4 & 71.9 & 69.9 & 56.4 & 46.8 & 63.4 \\
LLaMA3.1-70B-Instruct & 72.4 & 34.1 &  72.5 & 47.6 &  41.0 & 36.8 & 50.7 \\
Qwen2.5-72B-Instruct  & 72.6 & 45.3 & 72.2 & 52.4 &  41.9 & 43.1 & 54.6 \\
Qwen2.5-Math-72B-Instruct & 73.6 & 41.0 &  68.6 & 48.5 & 28.6 & 27.3 & 47.9 \\
GPT-4o & 69.7 & 45.9 & 72.1 & 57.3 & 50.5 & 53.4 & 58.2 \\
\midrule
\multicolumn{8}{l}{\emph{\quad \textbf{Our Critique Models}}} \\
DeepCritic-7B-SFT & 67.1 & 48.0 & 59.2 & 61.2 & 46.0 & 43.0 & 54.1 \\
DeepCritic-7B-RL-Numina & 77.2 & 55.9 & 70.7 & 65.9 & 57.6 & 53.5 & 63.5 \\
DeepCritic-7B-RL-PRM800K & 77.3 & \textbf{60.1} & \textbf{74.0} & \textbf{72.9} & \textbf{60.9} & 
\textbf{57.2} & \textbf{67.1} \\
\bottomrule
\end{tabular}
\end{table*}

The overall results on all benchmarks are displayed in Table~\ref{tab: main results}. We put the detailed results of separate accuracy on both incorrect and correct solutions in Appendix~\ref{appendix: detailed results on all benchmarks}.

First, we can observe that existing instruct models, especially those of small sizes, exhibit very limited critique capabilities, as reflected in their poor judgment performance. As the model size increases, the corresponding critique capability also increases. Second, improvements in the model's reasoning ability have a positive impact on its critique capability. This is reflected in that the strong reasoning abilities of the DeepSeek-R1-Distill models obtained in other domains can be transferred to the critique task and bring substantial performance gains. Third, \textbf{our seed critique model DeepCritic-7B-SFT, trained on 4.5K carefully curated deliberate critique data, achieves a 20-point F1 score improvement (34.1$\rightarrow$54.1) over the corresponding base model Qwen2.5-7B-Instruct}. Its overall performance is even comparable to that of Qwen2.5-72B-Instruct. This demonstrates the high quality of our constructed seed critique data and validates our motivation that teaching LLMs to perform deliberate critiquing can indeed lead to significant performance improvements.

Regarding the RL performance, we can see that \textbf{RL with only 14.2K automatically constructed data (i.e., DeepCritic-7B-RL-Numina) can effectively boost the model's critique performance from 54.1 to 63.5}. This validates the potential of automatically constructing supervision data and paves the promising way for the automated scalable oversight. Furthermore, when trained with larger scale RL data with higher quality, \textbf{the resulted model DeepCritic-7B-RL-PRM800K outperforms all baselines, including GPT-4o and two same-sized DeepSeek-R1-Distill models, in 5 out of 6 evaluation sets and achieves the best overall performance}. All in all, the above experimental results demonstrate that our proposed two-stage training paradigm is highly effective in enhancing the critique and verification capabilities of LLMs.

\section{Test-Time Scaling Results}
In this section, we explore the test-time scaling properties within the critique framework, from the perspectives of both critics and generators. In the following experiments, we choose our most powerful critique model so far DeepCritic-7B-RL-PRM800K as the target model, and refer to it as \textbf{DeepCritic-7B-RL} for brevity. We take the base model Qwen2.5-7B-Instruct along with the strongest baseline DeepSeek-R1-Distill-Qwen-7B (abbreviated to DS-R1-Distill-Qwen-7B) for comparison. 

\subsection{Test-Time Scaling Results of Critics}
\label{subsec: majority voting}
\label{subsec: results of majority voting}

\begin{figure*}[t]
  \centering
  \subfigure[Results on MR-GSM8K]{\includegraphics[width=0.315\textwidth]{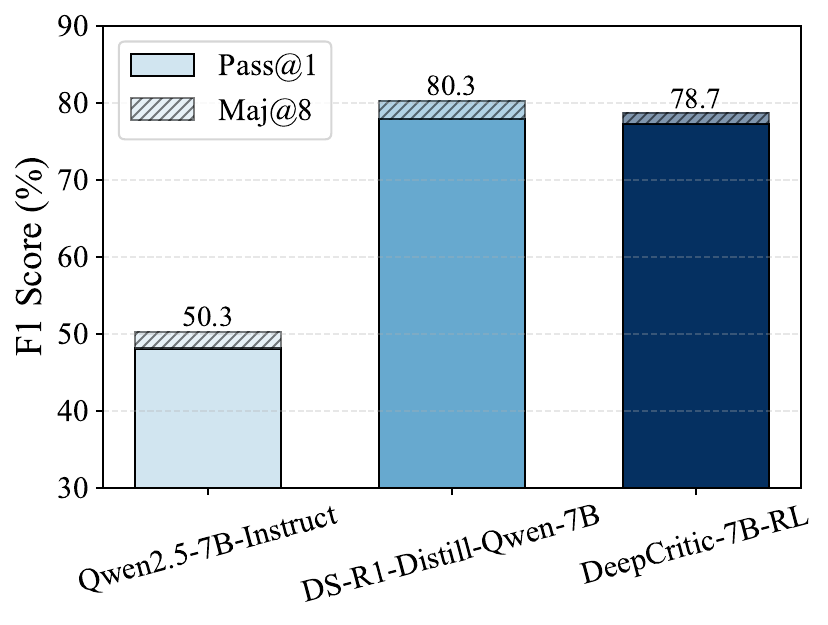}
  }
  \subfigure[Results on PRM800K]{\includegraphics[width=0.315\textwidth]{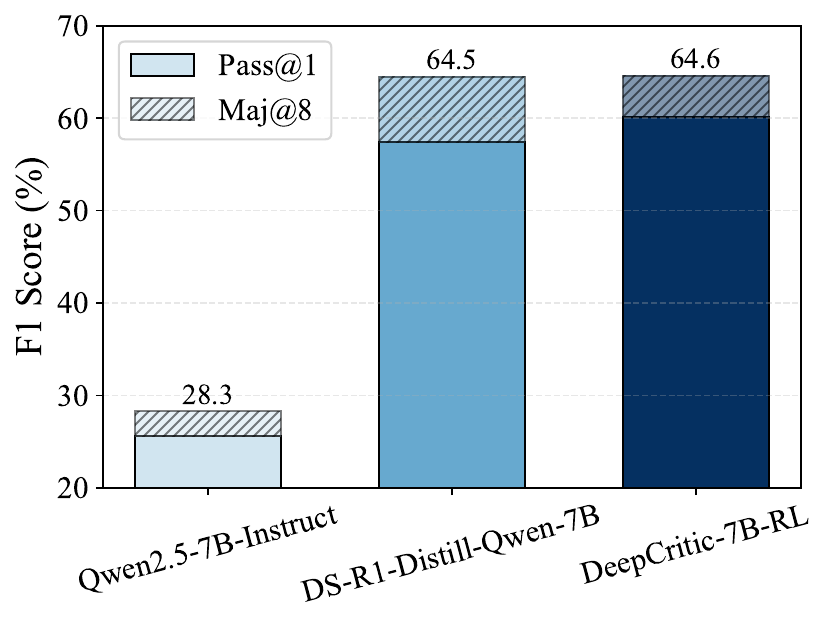}
  }
  \subfigure[Results on PB-GSM8K]{
    \includegraphics[width=0.315\textwidth]{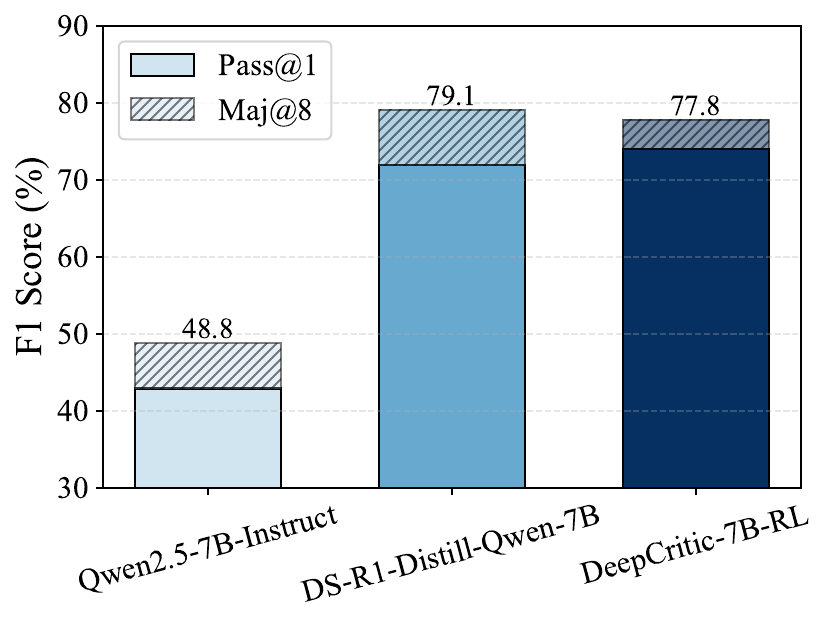}
  }
  \subfigure[Results on PB-MATH]{
    \includegraphics[width=0.315\textwidth]{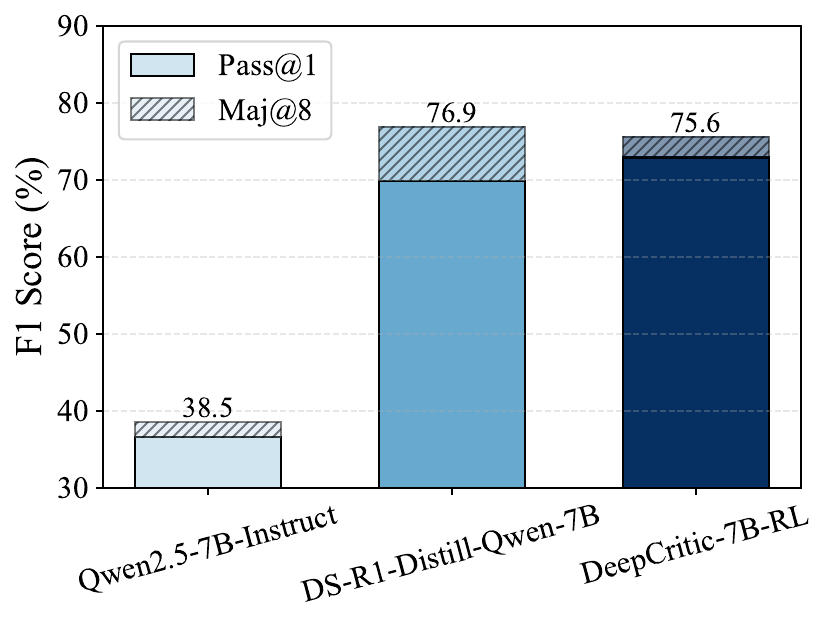}
  }
  \subfigure[Results on PB-OlympiadBench]{
    \includegraphics[width=0.315\textwidth]{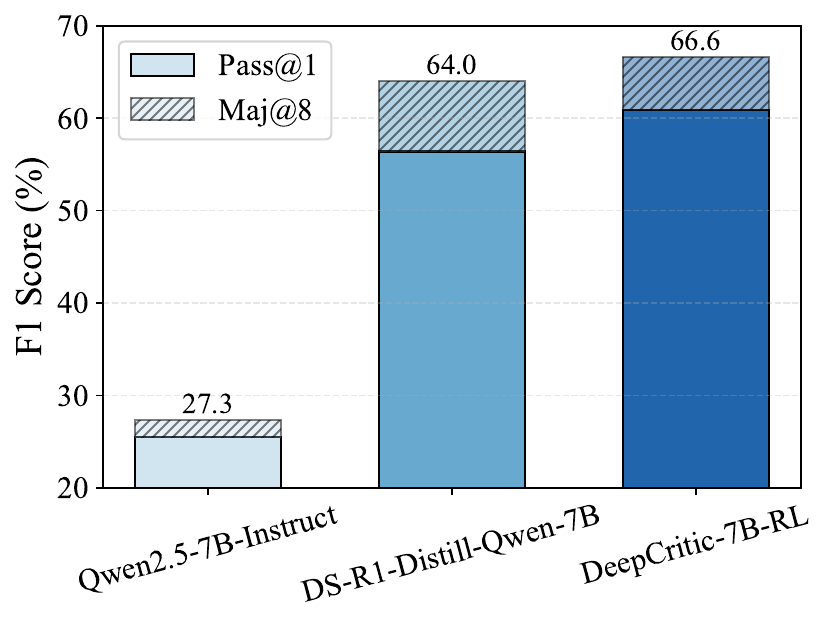}
  }
  \subfigure[Results on PB-Omni-Math]{
    \includegraphics[width=0.315\textwidth]{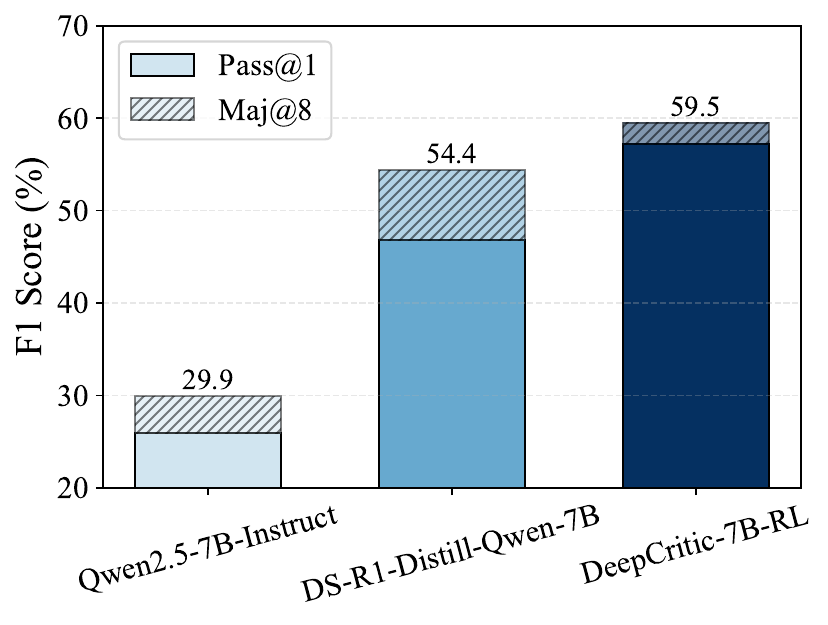}
  }
  \caption{Majority voting results (Maj@8) of each model across all benchmarks. Pass@1 results are from Table~\ref{tab: main results}. ``PB'' denotes ProcessBench. 
  }
  \label{fig: results of majority voting}
  \vskip -0.1in
\end{figure*}

Here, we investigate the effectiveness of the majority voting practice~\citep{self-consistency} in enhancing the critique performance. For each critique model, the final judgment on each input is the majority voting result over eight samplings (Maj@8). We put the comparison results between Maj@8 and Pass@1 in Figure~\ref{fig: results of majority voting}. As shown in Figure~\ref{fig: results of majority voting}, the majority voting practice improves performance across all models. The Maj@8 results of our critique model outperform DeepSeek-R1-Distill-Qwen-7B in half of the settings, and the average F1 score of our model (70.5) is also higher than that of DeepSeek-R1-Distill-Qwen-7B (69.9), demonstrating the good test-time scaling property of our critique model.

\subsection{Test-Time Scaling Results of Generators}
Critics can also be used to improve the performance of LLM generators via scaling the test-time compute of generators. On the one hand, similar to the role of PRMs, critics can act as verifiers to assess whether the responses sampled by the generator are correct. By filtering out identified incorrect responses, more accurate majority voting results of solutions can be achieved. On the other hand, the generator can revise potentially erroneous responses based on the feedback from the critic, thereby arriving at the correct answer through refinement. In the following, we delve into above two aspects separately. We select two generators of different sizes for experiments: Qwen2.5-7B-Instruct and Qwen2.5-72B-Instruct. We conduct evaluations on MATH500 and AIME2024-2025. We use \texttt{Math-Verify}\footnote{\url{https://github.com/huggingface/Math-Verify}} as the rule-based verifier to determine whether the predicted answer matches the ground truth answer.

\subsubsection{Results of Verified Majority Voting}

\begin{figure*}[t]
\centering
\begin{tabular}{c}
\includegraphics[width=0.98\linewidth]{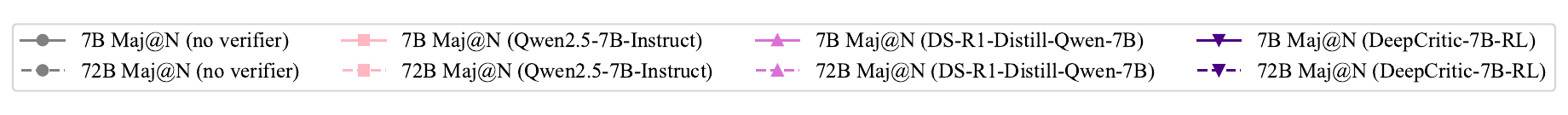}
\vspace{-1.0em}
\end{tabular}

\subfigure[Results on MATH500]{\includegraphics[width=0.48\textwidth]{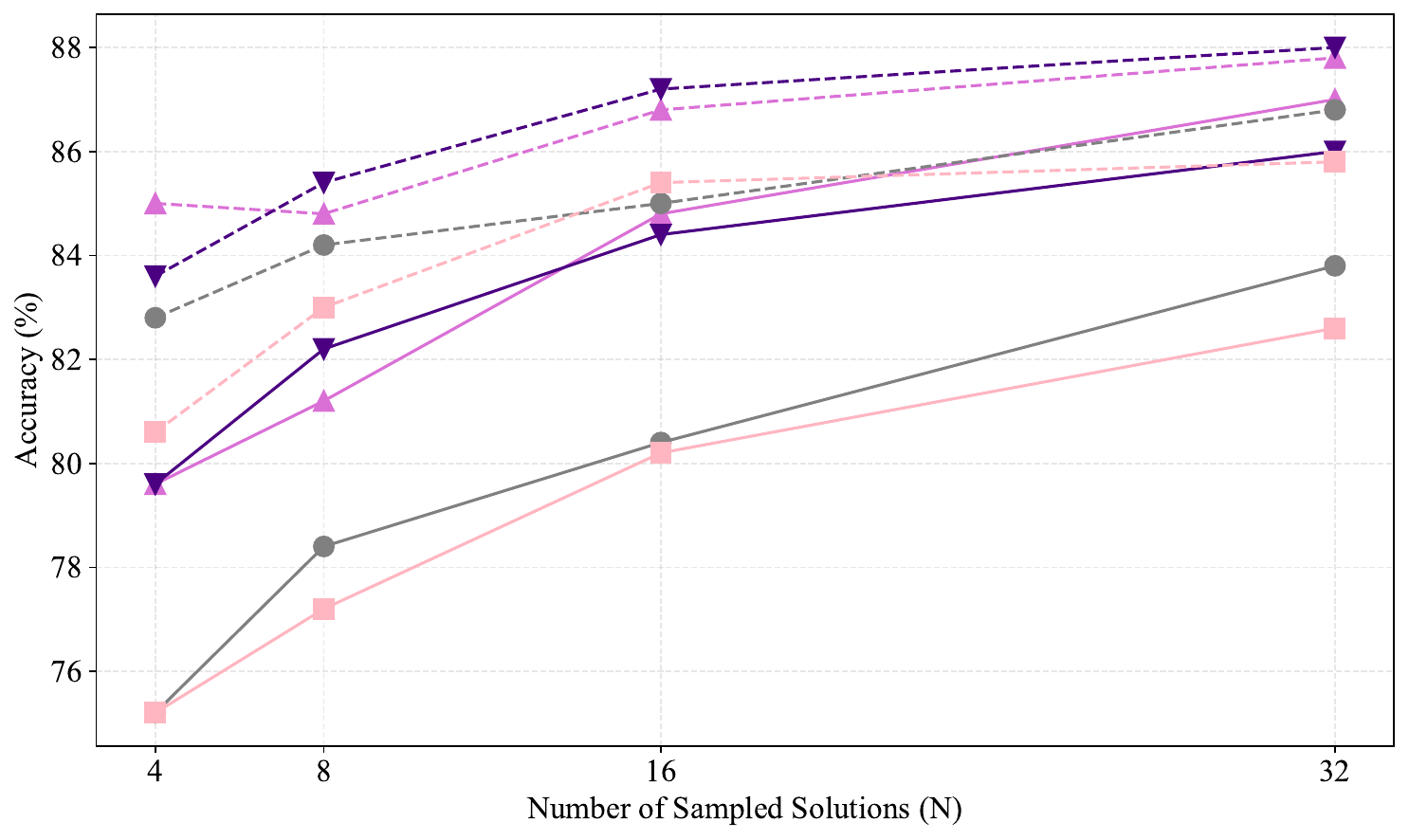}
  }
  \subfigure[Results on AIME24-25]{\includegraphics[width=0.48\textwidth]{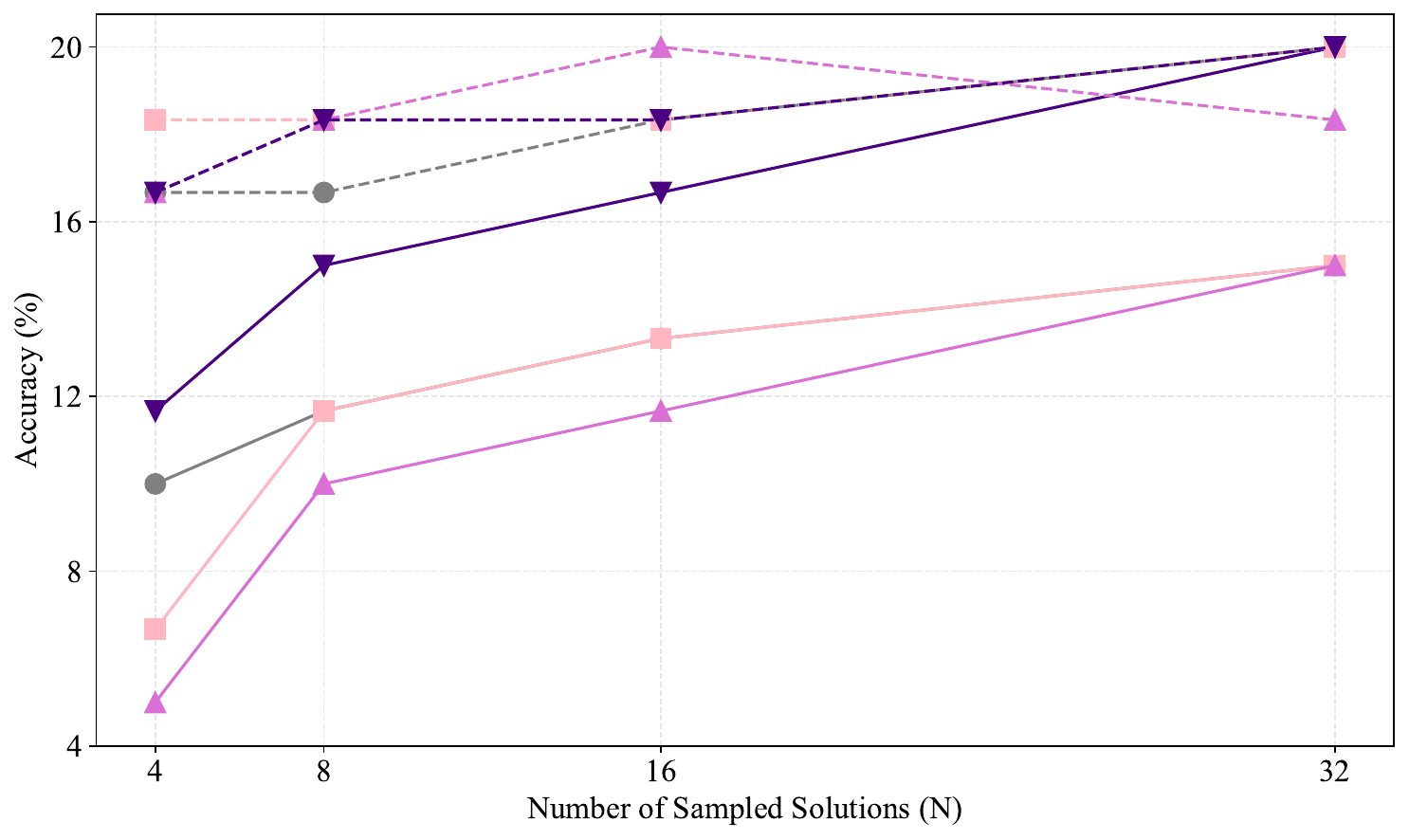}
  }

\caption{Verified majority voting results of Qwen2.5-7B/72B-Instruct on MATH500 and AIME2024-2025 by taking different models as verifiers.}
\label{fig: verified majority voting results}
\end{figure*}

We put the majority voting results under different numbers of sampled solutions from the generators, filtered taking the critique model as the verifier,\footnote{If all candidate solutions are identified as incorrect by the critique model, we fall back to perform majority voting over the original solutions.} in Figure~\ref{fig: verified majority voting results}. The sampling temperature for generators is set to 1.0.\footnote{In this setting and the following refinement experiments, to ensure that the generators produce responses in a required step-by-step format—enabling subsequent critiques by the critique model—we adopt a system prompt (refer to Appendix~\ref{appendix: refinement experiments settings}) different from the original one used by Qwen2.5 models. As a result, the evaluation results in Figure~\ref{fig: verified majority voting results} and Table~\ref{tab: refinement results} may differ from the original results.} We observe that when the critique model performs poorly (e.g., Qwen2.5-7B-Instruct), using it as the verifier in majority voting can be counterproductive. In contrast, our critique model can yield greater improvements to generators' majority voting performance in most sampling settings compared to baselines.

\subsubsection{Results of Critique-Based Refinement}

\begin{table*}[t]
\caption{Results of critique-based refinement. ``w$\rightarrow$c'' represents the proportion of cases where the model initially produces a wrong solution but arrives at the correct answer after judgment and refinement, and ``c$\rightarrow$w'' represents the the ratio of cases where correct solutions turns into incorrect one after judgment and refinement. ``Acc.'' represents the average accuracy on all testing samples. ``*'' denotes that the refinement results are biased due to DS-R1-Distill-Qwen-7B directly producing the correct answers during critique.
}
\label{tab: refinement results}
\centering
\small
\setlength{\tabcolsep}{2.2pt}
\begin{tabular}{lcccccccccccc}
\toprule
\multirow{4}{*}{Critique Model} & \multicolumn{6}{c}{Qwen2.5-7B-Instruct} & \multicolumn{6}{c}{Qwen2.5-72B-Instruct} \\
\cmidrule(lr){2-7}
\cmidrule(lr){8-13}
 & \multicolumn{3}{c}{MATH500} & \multicolumn{3}{c}{AIME24-25} & \multicolumn{3}{c}{MATH500} & \multicolumn{3}{c}{AIME24-25} \\
\cmidrule(lr){2-4}
\cmidrule(lr){5-7}
\cmidrule(lr){8-10}
\cmidrule(lr){11-13}
& w$\rightarrow$c & c$\rightarrow$w & Acc. & w$\rightarrow$c & c$\rightarrow$w & Acc. & w$\rightarrow$c & c$\rightarrow$w & Acc. & w$\rightarrow$c & c$\rightarrow$w & Acc. \\ 
\midrule
 \multicolumn{13}{l}{\emph{\quad \textbf{before refinement}}} \\
& --- &--- & 74.00 & --- & --- & \phantom{0}6.67 & --- & --- & 77.00 & --- & --- & 11.67 \\
\midrule
 \multicolumn{13}{l}{\emph{\quad \textbf{after refinement}}} \\
Qwen2.5-7B-Instruct & \phantom{0}0.80 & \phantom{0}2.60 & 72.20 & \phantom{0}1.67 & \phantom{0}0.00 & \phantom{0}8.33 & \phantom{0}1.60 & \phantom{0}2.40 & 76.20 & \phantom{0}1.67 & \phantom{0}0.00 & 13.33\\
DeepCritic-7B-RL & \phantom{0}4.60 & \phantom{0}1.40 & \textbf{77.20} & \phantom{0}5.00 & \phantom{0}0.00 & \textbf{11.67} & \phantom{0}7.00 & \phantom{0}2.00 & \textbf{82.00} &  \phantom{0}5.00 & \phantom{0}1.67 & \textbf{15.00}  \\
\midrule
 \multicolumn{13}{l}{\emph{\quad \textbf{after refinement (answer leakage)}}} \\
DS-R1-Distill-Qwen-7B* & \phantom{0}7.20 & \phantom{0}1.20 & 80.00 & \phantom{0}8.33 & \phantom{0}0.00 & 15.00  & \phantom{0}7.40 & \phantom{0}1.00 & 83.40 & \phantom{0}3.33 & \phantom{0}0.00 & 15.00 \\
\bottomrule
\end{tabular}
\end{table*}

In this setting, we first prompt generators to produce step-by-step solutions for each problem. Then, we leverage each critic to critique the solutions and prompt the generators to revise those deemed incorrect, based on the critic’s feedback. We use greedy decoding for the generators for determinism. In experiments, we observe that DeepSeek-R1-Distill-Qwen-7B frequently continues critiquing until the end of the solution and produces the correct answer, even though we explicitly instruct the model in the system prompt to stop after identifying the first incorrect step. This issue can cause the refinement results to be biased and greatly influenced by DeekSeek-R1-Distill-Qwen-7B's own problem-solving capability. Therefore, we present its results independently for reference. The refinement results are shown in Table~\ref{tab: refinement results}. We can see that our critique model can effectively assist the generators in correcting errors by providing more detailed feedback, leading to improved performance of the generators. Notably, our 7B critique model is also capable of supervising and correcting the outputs of a 72B generator, demonstrating a potential of weak-to-strong supervision~\cite{weak-to-strong}.

\section{Conclusion}
In this work, we propose an effective pipeline to enhance the math critique ability of LLMs. We first carefully construct 4.5K long-form critiques incorporating multi-perspective verification and meta-critiquing. These serve as seed data for supervised fine-tuning, enabling the target model to acquire an initial ability of deliberately critiquing. We then further enhance the critique capability of the model via reinforcement learning. The deep critique model we developed demonstrates superior performance across a range of error identification benchmarks, and exhibits promising potential in supervising and improving the reasoning capabilities of LLM generators that are even more capable than itself. We hope our work provides valuable insights to advance future research in deliberate reasoning and scalable oversight.

\bibliography{neurips_2024}

\begin{thebibliography}{56}
\providecommand{\natexlab}[1]{#1}
\providecommand{\url}[1]{\texttt{#1}}
\expandafter\ifx\csname urlstyle\endcsname\relax
  \providecommand{\doi}[1]{doi: #1}\else
  \providecommand{\doi}{doi: \begingroup \urlstyle{rm}\Url}\fi

\bibitem[Bai et~al.(2022)Bai, Jones, Ndousse, Askell, Chen, DasSarma, Drain, Fort, Ganguli, Henighan, et~al.]{hh-rlhf}
Yuntao Bai, Andy Jones, Kamal Ndousse, Amanda Askell, Anna Chen, Nova DasSarma, Dawn Drain, Stanislav Fort, Deep Ganguli, Tom Henighan, et~al.
\newblock Training a helpful and harmless assistant with reinforcement learning from human feedback.
\newblock \emph{arXiv preprint arXiv:2204.05862}, 2022.

\bibitem[Bowman et~al.(2022)Bowman, Hyun, Perez, Chen, Pettit, Heiner, Luko{\v{s}}i{\=u}t{\.e}, Askell, Jones, Chen, et~al.]{measureing_scalable_oversight}
Samuel~R Bowman, Jeeyoon Hyun, Ethan Perez, Edwin Chen, Craig Pettit, Scott Heiner, Kamil{\.e} Luko{\v{s}}i{\=u}t{\.e}, Amanda Askell, Andy Jones, Anna Chen, et~al.
\newblock Measuring progress on scalable oversight for large language models.
\newblock \emph{arXiv preprint arXiv:2211.03540}, 2022.

\bibitem[Burns et~al.(2024)Burns, Izmailov, Kirchner, Baker, Gao, Aschenbrenner, Chen, Ecoffet, Joglekar, Leike, et~al.]{weak-to-strong}
Collin Burns, Pavel Izmailov, Jan~Hendrik Kirchner, Bowen Baker, Leo Gao, Leopold Aschenbrenner, Yining Chen, Adrien Ecoffet, Manas Joglekar, Jan Leike, et~al.
\newblock Weak-to-strong generalization: eliciting strong capabilities with weak supervision.
\newblock In \emph{Proceedings of the 41st International Conference on Machine Learning}, pages 4971--5012, 2024.

\bibitem[Cobbe et~al.(2021)Cobbe, Kosaraju, Bavarian, Chen, Jun, Kaiser, Plappert, Tworek, Hilton, Nakano, et~al.]{gsm8k}
Karl Cobbe, Vineet Kosaraju, Mohammad Bavarian, Mark Chen, Heewoo Jun, Lukasz Kaiser, Matthias Plappert, Jerry Tworek, Jacob Hilton, Reiichiro Nakano, et~al.
\newblock Training verifiers to solve math word problems.
\newblock \emph{arXiv preprint arXiv:2110.14168}, 2021.

\bibitem[Cui et~al.(2025)Cui, Yuan, Wang, Wang, Li, He, Fan, Yu, Xu, Chen, et~al.]{prime}
Ganqu Cui, Lifan Yuan, Zefan Wang, Hanbin Wang, Wendi Li, Bingxiang He, Yuchen Fan, Tianyu Yu, Qixin Xu, Weize Chen, et~al.
\newblock Process reinforcement through implicit rewards.
\newblock \emph{arXiv preprint arXiv:2502.01456}, 2025.

\bibitem[{DeepSeek}(2025)]{r1}
{DeepSeek}.
\newblock Deepseek-r1: Incentivizing reasoning capability in llms via reinforcement learning, 1 2025.
\newblock URL \url{https://github.com/deepseek-ai/DeepSeek-R1/blob/main/DeepSeek_R1.pdf}.

\bibitem[Gao et~al.(2024{\natexlab{a}})Gao, Cai, Xu, Wang, Zheng, Lin, Lu, Liu, Zhou, Xiao, et~al.]{llm-critic-catch-math-bugs}
Bofei Gao, Zefan Cai, Runxin Xu, Peiyi Wang, Ce~Zheng, Runji Lin, Keming Lu, Dayiheng Liu, Chang Zhou, Wen Xiao, et~al.
\newblock Llm critics help catch bugs in mathematics: Towards a better mathematical verifier with natural language feedback.
\newblock \emph{arXiv preprint arXiv:2406.14024}, 2024{\natexlab{a}}.

\bibitem[Gao et~al.(2024{\natexlab{b}})Gao, Song, Yang, Cai, Miao, Dong, Li, Ma, Chen, Xu, et~al.]{omni-math}
Bofei Gao, Feifan Song, Zhe Yang, Zefan Cai, Yibo Miao, Qingxiu Dong, Lei Li, Chenghao Ma, Liang Chen, Runxin Xu, et~al.
\newblock Omni-math: A universal olympiad level mathematic benchmark for large language models.
\newblock \emph{arXiv preprint arXiv:2410.07985}, 2024{\natexlab{b}}.

\bibitem[Geva et~al.(2021)Geva, Khashabi, Segal, Khot, Roth, and Berant]{strategyqa}
Mor Geva, Daniel Khashabi, Elad Segal, Tushar Khot, Dan Roth, and Jonathan Berant.
\newblock Did aristotle use a laptop? a question answering benchmark with implicit reasoning strategies.
\newblock \emph{Transactions of the Association for Computational Linguistics}, 9:\penalty0 346--361, 2021.

\bibitem[Gu et~al.(2024)Gu, Jiang, Shi, Tan, Zhai, Xu, Li, Shen, Ma, Liu, et~al.]{survey-on-llm-as-a-judge}
Jiawei Gu, Xuhui Jiang, Zhichao Shi, Hexiang Tan, Xuehao Zhai, Chengjin Xu, Wei Li, Yinghan Shen, Shengjie Ma, Honghao Liu, et~al.
\newblock A survey on llm-as-a-judge.
\newblock \emph{arXiv preprint arXiv:2411.15594}, 2024.

\bibitem[He et~al.(2024)He, Luo, Bai, Hu, Thai, Shen, Hu, Han, Huang, Zhang, et~al.]{olympiadbench}
Chaoqun He, Renjie Luo, Yuzhuo Bai, Shengding Hu, Zhen Thai, Junhao Shen, Jinyi Hu, Xu~Han, Yujie Huang, Yuxiang Zhang, et~al.
\newblock Olympiadbench: A challenging benchmark for promoting agi with olympiad-level bilingual multimodal scientific problems.
\newblock In \emph{Proceedings of the 62nd Annual Meeting of the Association for Computational Linguistics (Volume 1: Long Papers)}, pages 3828--3850, 2024.

\bibitem[Hendrycks et~al.(2021)Hendrycks, Burns, Kadavath, Arora, Basart, Tang, Song, and Steinhardt]{math}
Dan Hendrycks, Collin Burns, Saurav Kadavath, Akul Arora, Steven Basart, Eric Tang, Dawn Song, and Jacob Steinhardt.
\newblock Measuring mathematical problem solving with the {MATH} dataset.
\newblock In \emph{Thirty-fifth Conference on Neural Information Processing Systems Datasets and Benchmarks Track (Round 2)}, 2021.
\newblock URL \url{https://openreview.net/forum?id=7Bywt2mQsCe}.

\bibitem[Hu et~al.(2025)Hu, Zhang, Han, Jiang, Zhang, and Shum]{open-reasoner-zero}
Jingcheng Hu, Yinmin Zhang, Qi~Han, Daxin Jiang, Xiangyu Zhang, and Heung-Yeung Shum.
\newblock Open-reasoner-zero: An open source approach to scaling up reinforcement learning on the base model, 2025.
\newblock URL \url{https://arxiv.org/abs/2503.24290}.

\bibitem[Hui et~al.(2024)Hui, Yang, Cui, Yang, Liu, Zhang, Liu, Zhang, Yu, Lu, et~al.]{qwen2.5-coder}
Binyuan Hui, Jian Yang, Zeyu Cui, Jiaxi Yang, Dayiheng Liu, Lei Zhang, Tianyu Liu, Jiajun Zhang, Bowen Yu, Keming Lu, et~al.
\newblock Qwen2. 5-coder technical report.
\newblock \emph{arXiv preprint arXiv:2409.12186}, 2024.

\bibitem[Hurst et~al.(2024)Hurst, Lerer, Goucher, Perelman, Ramesh, Clark, Ostrow, Welihinda, Hayes, Radford, et~al.]{gpt4o}
Aaron Hurst, Adam Lerer, Adam~P Goucher, Adam Perelman, Aditya Ramesh, Aidan Clark, AJ~Ostrow, Akila Welihinda, Alan Hayes, Alec Radford, et~al.
\newblock Gpt-4o system card.
\newblock \emph{arXiv preprint arXiv:2410.21276}, 2024.

\bibitem[Jimenez et~al.(2024)Jimenez, Yang, Wettig, Yao, Pei, Press, and Narasimhan]{swebench}
Carlos~E Jimenez, John Yang, Alexander Wettig, Shunyu Yao, Kexin Pei, Ofir Press, and Karthik~R Narasimhan.
\newblock {SWE}-bench: Can language models resolve real-world github issues?
\newblock In \emph{The Twelfth International Conference on Learning Representations}, 2024.
\newblock URL \url{https://openreview.net/forum?id=VTF8yNQM66}.

\bibitem[Ke et~al.(2024)Ke, Wen, Feng, Liu, Lei, Cheng, Wang, Zeng, Dong, Wang, Tang, and Huang]{critiquellm}
Pei Ke, Bosi Wen, Andrew Feng, Xiao Liu, Xuanyu Lei, Jiale Cheng, Shengyuan Wang, Aohan Zeng, Yuxiao Dong, Hongning Wang, Jie Tang, and Minlie Huang.
\newblock {C}ritique{LLM}: Towards an informative critique generation model for evaluation of large language model generation.
\newblock In Lun-Wei Ku, Andre Martins, and Vivek Srikumar, editors, \emph{Proceedings of the 62nd Annual Meeting of the Association for Computational Linguistics (Volume 1: Long Papers)}, pages 13034--13054, Bangkok, Thailand, August 2024. Association for Computational Linguistics.
\newblock \doi{10.18653/v1/2024.acl-long.704}.
\newblock URL \url{https://aclanthology.org/2024.acl-long.704/}.

\bibitem[Lai et~al.(2024)Lai, Tian, Chen, Yang, Peng, and Jia]{step-dpo}
Xin Lai, Zhuotao Tian, Yukang Chen, Senqiao Yang, Xiangru Peng, and Jiaya Jia.
\newblock Step-dpo: Step-wise preference optimization for long-chain reasoning of llms.
\newblock \emph{arXiv preprint arXiv:2406.18629}, 2024.

\bibitem[Lan et~al.(2024)Lan, Zhang, Xu, Huang, Lin, Chen, and Mao]{criticeval}
Tian Lan, Wenwei Zhang, Chen Xu, Heyan Huang, Dahua Lin, Kai Chen, and Xian-Ling Mao.
\newblock Criticeval: Evaluating large-scale language model as critic.
\newblock \emph{Advances in Neural Information Processing Systems}, 37:\penalty0 66907--66960, 2024.

\bibitem[Li et~al.(2024)Li, Beeching, Tunstall, Lipkin, Soletskyi, Huang, Rasul, Yu, Jiang, Shen, et~al.]{numinamath}
Jia Li, Edward Beeching, Lewis Tunstall, Ben Lipkin, Roman Soletskyi, Shengyi Huang, Kashif Rasul, Longhui Yu, Albert~Q Jiang, Ziju Shen, et~al.
\newblock Numinamath: The largest public dataset in ai4maths with 860k pairs of competition math problems and solutions.
\newblock \emph{Hugging Face repository}, 13, 2024.

\bibitem[Lightman et~al.(2023)Lightman, Kosaraju, Burda, Edwards, Baker, Lee, Leike, Schulman, Sutskever, and Cobbe]{verify_step_by_step}
Hunter Lightman, Vineet Kosaraju, Yuri Burda, Harrison Edwards, Bowen Baker, Teddy Lee, Jan Leike, John Schulman, Ilya Sutskever, and Karl Cobbe.
\newblock Let's verify step by step.
\newblock In \emph{The Twelfth International Conference on Learning Representations}, 2023.

\bibitem[Lin et~al.(2024)Lin, Gou, Liang, Luo, Liu, and Yang]{criticbench}
Zicheng Lin, Zhibin Gou, Tian Liang, Ruilin Luo, Haowei Liu, and Yujiu Yang.
\newblock Criticbench: Benchmarking llms for critique-correct reasoning.
\newblock In \emph{Findings of the Association for Computational Linguistics ACL 2024}, pages 1552--1587, 2024.

\bibitem[Liu et~al.(2024)Liu, Feng, Xue, Wang, Wu, Lu, Zhao, Deng, Zhang, Ruan, et~al.]{deepseek-v3}
Aixin Liu, Bei Feng, Bing Xue, Bingxuan Wang, Bochao Wu, Chengda Lu, Chenggang Zhao, Chengqi Deng, Chenyu Zhang, Chong Ruan, et~al.
\newblock Deepseek-v3 technical report.
\newblock \emph{arXiv preprint arXiv:2412.19437}, 2024.

\bibitem[Luo et~al.(2023)Luo, Lin, Liu, Shu, Zhu, Shang, and Meng]{Critique_ability}
Liangchen Luo, Zi~Lin, Yinxiao Liu, Lei Shu, Yun Zhu, Jingbo Shang, and Lei Meng.
\newblock Critique ability of large language models.
\newblock \emph{arXiv preprint arXiv:2310.04815}, 2023.

\bibitem[Luo et~al.(2024)Luo, Rechardt, Sun, Nejad, Y{\'a}{\~n}ez, Yilmaz, Lee, Cohen, Borghesani, Pashkov, et~al.]{llm_predict_neuroscience}
Xiaoliang Luo, Akilles Rechardt, Guangzhi Sun, Kevin~K Nejad, Felipe Y{\'a}{\~n}ez, Bati Yilmaz, Kangjoo Lee, Alexandra~O Cohen, Valentina Borghesani, Anton Pashkov, et~al.
\newblock Large language models surpass human experts in predicting neuroscience results.
\newblock \emph{Nature human behaviour}, pages 1--11, 2024.

\bibitem[Madaan et~al.(2023)Madaan, Tandon, Gupta, Hallinan, Gao, Wiegreffe, Alon, Dziri, Prabhumoye, Yang, et~al.]{self-refine}
Aman Madaan, Niket Tandon, Prakhar Gupta, Skyler Hallinan, Luyu Gao, Sarah Wiegreffe, Uri Alon, Nouha Dziri, Shrimai Prabhumoye, Yiming Yang, et~al.
\newblock Self-refine: Iterative refinement with self-feedback.
\newblock \emph{Advances in Neural Information Processing Systems}, 36:\penalty0 46534--46594, 2023.

\bibitem[McAleese et~al.(2024)McAleese, Pokorny, Uribe, Nitishinskaya, Trebacz, and Leike]{llm-critics-help-catch-llm-bugs}
Nat McAleese, Rai~Michael Pokorny, Juan Felipe~Ceron Uribe, Evgenia Nitishinskaya, Maja Trebacz, and Jan Leike.
\newblock Llm critics help catch llm bugs.
\newblock \emph{arXiv preprint arXiv:2407.00215}, 2024.

\bibitem[MetaAI(2024)]{llama3.1}
MetaAI.
\newblock Introducing llama 3.1: Our most capable models to date.
\newblock \url{https://ai.meta.com/blog/meta-llama-3-1/}, 2024.

\bibitem[OpenAI(2023)]{gpt4}
OpenAI.
\newblock Gpt-4 technical report.
\newblock \emph{arXiv}, pages 2303--08774, 2023.

\bibitem[OpenAI(2024)]{o1}
OpenAI.
\newblock Learning to reason with llms, 2024.
\newblock URL \url{https://openai.com/index/learning-to-reason-with-llms}.

\bibitem[Ouyang et~al.(2022)Ouyang, Wu, Jiang, Almeida, Wainwright, Mishkin, Zhang, Agarwal, Slama, Ray, et~al.]{instructGPT}
Long Ouyang, Jeffrey Wu, Xu~Jiang, Diogo Almeida, Carroll Wainwright, Pamela Mishkin, Chong Zhang, Sandhini Agarwal, Katarina Slama, Alex Ray, et~al.
\newblock Training language models to follow instructions with human feedback.
\newblock \emph{Advances in Neural Information Processing Systems}, 35:\penalty0 27730--27744, 2022.

\bibitem[{Qwen Team}(2024)]{qwen2.5}
{Qwen Team}.
\newblock Qwen2. 5: A party of foundation models.
\newblock \emph{Qwen (Sept. 2024). url: https://qwenlm. github. io/blog/qwen2}, 5, 2024.

\bibitem[Rein et~al.(2023)Rein, Hou, Stickland, Petty, Pang, Dirani, Michael, and Bowman]{gpqa}
David Rein, Betty~Li Hou, Asa~Cooper Stickland, Jackson Petty, Richard~Yuanzhe Pang, Julien Dirani, Julian Michael, and Samuel~R Bowman.
\newblock Gpqa: A graduate-level google-proof q\&a benchmark.
\newblock \emph{arXiv preprint arXiv:2311.12022}, 2023.

\bibitem[Roziere et~al.(2023)Roziere, Gehring, Gloeckle, Sootla, Gat, Tan, Adi, Liu, Sauvestre, Remez, et~al.]{codellama}
Baptiste Roziere, Jonas Gehring, Fabian Gloeckle, Sten Sootla, Itai Gat, Xiaoqing~Ellen Tan, Yossi Adi, Jingyu Liu, Romain Sauvestre, Tal Remez, et~al.
\newblock Code llama: Open foundation models for code.
\newblock \emph{arXiv preprint arXiv:2308.12950}, 2023.

\bibitem[Saunders et~al.(2022)Saunders, Yeh, Wu, Bills, Ouyang, Ward, and Leike]{scalable-oversight}
William Saunders, Catherine Yeh, Jeff Wu, Steven Bills, Long Ouyang, Jonathan Ward, and Jan Leike.
\newblock Self-critiquing models for assisting human evaluators.
\newblock \emph{arXiv preprint arXiv:2206.05802}, 2022.

\bibitem[Shao et~al.(2024)Shao, Wang, Zhu, Xu, Song, Bi, Zhang, Zhang, Li, Wu, et~al.]{deepseekmath}
Zhihong Shao, Peiyi Wang, Qihao Zhu, Runxin Xu, Junxiao Song, Xiao Bi, Haowei Zhang, Mingchuan Zhang, YK~Li, Y~Wu, et~al.
\newblock Deepseekmath: Pushing the limits of mathematical reasoning in open language models.
\newblock \emph{arXiv preprint arXiv:2402.03300}, 2024.

\bibitem[Sheng et~al.(2024)Sheng, Zhang, Ye, Wu, Zhang, Zhang, Peng, Lin, and Wu]{verl}
Guangming Sheng, Chi Zhang, Zilingfeng Ye, Xibin Wu, Wang Zhang, Ru~Zhang, Yanghua Peng, Haibin Lin, and Chuan Wu.
\newblock Hybridflow: A flexible and efficient rlhf framework.
\newblock \emph{arXiv preprint arXiv:2409.19256}, 2024.

\bibitem[Tang et~al.(2025)Tang, Li, Xiao, Ding, Sun, Wang, Liu, Huang, Liu, Yu, et~al.]{scrit}
Zhengyang Tang, Ziniu Li, Zhenyang Xiao, Tian Ding, Ruoyu Sun, Benyou Wang, Dayiheng Liu, Fei Huang, Tianyu Liu, Bowen Yu, et~al.
\newblock Enabling scalable oversight via self-evolving critic.
\newblock \emph{arXiv preprint arXiv:2501.05727}, 2025.

\bibitem[Uesato et~al.(2022)Uesato, Kushman, Kumar, Song, Siegel, Wang, Creswell, Irving, and Higgins]{process-and_outcome-based_feedback}
Jonathan Uesato, Nate Kushman, Ramana Kumar, Francis Song, Noah Siegel, Lisa Wang, Antonia Creswell, Geoffrey Irving, and Irina Higgins.
\newblock Solving math word problems with process-and outcome-based feedback.
\newblock \emph{arXiv preprint arXiv:2211.14275}, 2022.

\bibitem[Wang et~al.(2024)Wang, Li, Shao, Xu, Dai, Li, Chen, Wu, and Sui]{Math-shepherd}
Peiyi Wang, Lei Li, Zhihong Shao, Runxin Xu, Damai Dai, Yifei Li, Deli Chen, Yu~Wu, and Zhifang Sui.
\newblock Math-shepherd: Verify and reinforce llms step-by-step without human annotations.
\newblock In \emph{Proceedings of the 62nd Annual Meeting of the Association for Computational Linguistics (Volume 1: Long Papers)}, pages 9426--9439, 2024.

\bibitem[Wang et~al.(2023)Wang, Wei, Schuurmans, Le, Chi, Narang, Chowdhery, and Zhou]{self-consistency}
Xuezhi Wang, Jason Wei, Dale Schuurmans, Quoc~V Le, Ed~H. Chi, Sharan Narang, Aakanksha Chowdhery, and Denny Zhou.
\newblock Self-consistency improves chain of thought reasoning in language models.
\newblock In \emph{The Eleventh International Conference on Learning Representations}, 2023.
\newblock URL \url{https://openreview.net/forum?id=1PL1NIMMrw}.

\bibitem[Wu et~al.(2024{\natexlab{a}})Wu, Yuan, Golovneva, Xu, Tian, Jiao, Weston, and Sukhbaatar]{meta-rewarding}
Tianhao Wu, Weizhe Yuan, Olga Golovneva, Jing Xu, Yuandong Tian, Jiantao Jiao, Jason Weston, and Sainbayar Sukhbaatar.
\newblock Meta-rewarding language models: Self-improving alignment with llm-as-a-meta-judge.
\newblock \emph{arXiv preprint arXiv:2407.19594}, 2024{\natexlab{a}}.

\bibitem[Wu et~al.(2024{\natexlab{b}})Wu, Sun, Li, Welleck, and Yang]{compute-optimal-scaling}
Yangzhen Wu, Zhiqing Sun, Shanda Li, Sean Welleck, and Yiming Yang.
\newblock Scaling inference computation: Compute-optimal inference for problem-solving with language models.
\newblock In \emph{The 4th Workshop on Mathematical Reasoning and AI at NeurIPS'24}, 2024{\natexlab{b}}.
\newblock URL \url{https://openreview.net/forum?id=j7DZWSc8qu}.

\bibitem[Xi et~al.(2024)Xi, Yang, Huang, Tang, Li, Ding, He, Hong, Do, Zhan, et~al.]{enhancing-llm-via-critique}
Zhiheng Xi, Dingwen Yang, Jixuan Huang, Jiafu Tang, Guanyu Li, Yiwen Ding, Wei He, Boyang Hong, Shihan Do, Wenyu Zhan, et~al.
\newblock Enhancing llm reasoning via critique models with test-time and training-time supervision.
\newblock \emph{arXiv preprint arXiv:2411.16579}, 2024.

\bibitem[Xie et~al.(2025)Xie, Chen, Mao, Xu, Kong, et~al.]{ctrl}
Zhihui Xie, Liyu Chen, Weichao Mao, Jingjing Xu, Lingpeng Kong, et~al.
\newblock Teaching language models to critique via reinforcement learning.
\newblock \emph{arXiv preprint arXiv:2502.03492}, 2025.

\bibitem[Xiong et~al.(2024{\natexlab{a}})Xiong, Zhang, Jiang, and Zhang]{generative_prm}
Wei Xiong, Hanning Zhang, Nan Jiang, and Tong Zhang.
\newblock An implementation of generative prm, 2024{\natexlab{a}}.

\bibitem[Xiong et~al.(2024{\natexlab{b}})Xiong, Zhang, Jiang, and Zhang]{rlhflowmath}
Wei Xiong, Hanning Zhang, Nan Jiang, and Tong Zhang.
\newblock An implementation of generative prm.
\newblock \url{https://github.com/RLHFlow/RLHF-Reward-Modeling}, 2024{\natexlab{b}}.

\bibitem[Yang et~al.(2024{\natexlab{a}})Yang, Zhang, Hui, Gao, Yu, Li, Liu, Tu, Zhou, Lin, et~al.]{qwen2.5-math}
An~Yang, Beichen Zhang, Binyuan Hui, Bofei Gao, Bowen Yu, Chengpeng Li, Dayiheng Liu, Jianhong Tu, Jingren Zhou, Junyang Lin, et~al.
\newblock Qwen2. 5-math technical report: Toward mathematical expert model via self-improvement.
\newblock \emph{arXiv preprint arXiv:2409.12122}, 2024{\natexlab{a}}.

\bibitem[Yang et~al.(2024{\natexlab{b}})Yang, Shen, Shen, Yao, Liu, Gong, Lin, and Wen]{deceive_weak_models}
Wenkai Yang, Shiqi Shen, Guangyao Shen, Wei Yao, Yong Liu, Zhi Gong, Yankai Lin, and Ji-Rong Wen.
\newblock Super (ficial)-alignment: Strong models may deceive weak models in weak-to-strong generalization.
\newblock \emph{arXiv preprint arXiv:2406.11431}, 2024{\natexlab{b}}.

\bibitem[Yang et~al.(2025)Yang, Ma, Lin, and Wei]{Thinking-Optimal_Scaling}
Wenkai Yang, Shuming Ma, Yankai Lin, and Furu Wei.
\newblock Towards thinking-optimal scaling of test-time compute for llm reasoning.
\newblock \emph{arXiv preprint arXiv:2502.18080}, 2025.

\bibitem[Yu et~al.(2024)Yu, Jiang, Shi, YU, Liu, Zhang, Kwok, Li, Weller, and Liu]{metamath}
Longhui Yu, Weisen Jiang, Han Shi, Jincheng YU, Zhengying Liu, Yu~Zhang, James Kwok, Zhenguo Li, Adrian Weller, and Weiyang Liu.
\newblock Metamath: Bootstrap your own mathematical questions for large language models.
\newblock In \emph{The Twelfth International Conference on Learning Representations}, 2024.
\newblock URL \url{https://openreview.net/forum?id=N8N0hgNDRt}.

\bibitem[Yuan et~al.()Yuan, Pang, Cho, Li, Sukhbaatar, Xu, and Weston]{self-rewarding}
Weizhe Yuan, Richard~Yuanzhe Pang, Kyunghyun Cho, Xian Li, Sainbayar Sukhbaatar, Jing Xu, and Jason Weston.
\newblock Self-rewarding language models, 2024.
\newblock \emph{URL https://arxiv. org/abs/2401.10020}.

\bibitem[Zeng et~al.(2023)Zeng, Chen, Liu, Jiang, and Jia]{Mr-gsm8k}
Zhongshen Zeng, Pengguang Chen, Shu Liu, Haiyun Jiang, and Jiaya Jia.
\newblock Mr-gsm8k: A meta-reasoning benchmark for large language model evaluation.
\newblock \emph{arXiv preprint arXiv:2312.17080}, 2023.

\bibitem[Zheng et~al.(2024{\natexlab{a}})Zheng, Zhang, Zhang, Lin, Lu, Yu, Liu, Zhou, and Lin]{Processbench}
Chujie Zheng, Zhenru Zhang, Beichen Zhang, Runji Lin, Keming Lu, Bowen Yu, Dayiheng Liu, Jingren Zhou, and Junyang Lin.
\newblock Processbench: Identifying process errors in mathematical reasoning.
\newblock \emph{arXiv preprint arXiv:2412.06559}, 2024{\natexlab{a}}.

\bibitem[Zheng et~al.(2023)Zheng, Chiang, Sheng, Zhuang, Wu, Zhuang, Lin, Li, Li, Xing, et~al.]{llm-as-a-judge}
Lianmin Zheng, Wei-Lin Chiang, Ying Sheng, Siyuan Zhuang, Zhanghao Wu, Yonghao Zhuang, Zi~Lin, Zhuohan Li, Dacheng Li, Eric Xing, et~al.
\newblock Judging llm-as-a-judge with mt-bench and chatbot arena.
\newblock \emph{Advances in Neural Information Processing Systems}, 36:\penalty0 46595--46623, 2023.

\bibitem[Zheng et~al.(2024{\natexlab{b}})Zheng, Lou, Cao, Wen, Ji, Lin, Lu, Han, Zhang, and Sun]{critic-cot}
Xin Zheng, Jie Lou, Boxi Cao, Xueru Wen, Yuqiu Ji, Hongyu Lin, Yaojie Lu, Xianpei Han, Debing Zhang, and Le~Sun.
\newblock Critic-cot: Boosting the reasoning abilities of large language model via chain-of-thoughts critic.
\newblock \emph{arXiv preprint arXiv:2408.16326}, 2024{\natexlab{b}}.

\end{thebibliography}

\newpage
\appendix

\section{Prompt Templates for Critique Data Generation}
\label{appendix: prompt templates}
\begin{prompt}{Prompt Template for Initial Critique Generation}
\label{prompt: initial prompt generation}
You are a math expert and are tasked with evaluating the solution path for a mathematical problem.
\\
The solution is presented as a step-by-step chain of thought.
\\
Each step is separated with the index indicator ``Step i:'', with i indexed starting from 1.
\\
You are required to only critique the specific step carefully and comprehensively.
\\
You need to thoroughly consider the logical consistency of the specified step with the problem statement and previous steps, ensuring each step aligns with the overall and correct objective. 
\\
You should consider the cases where the steps are merely irrelevant transitions as correct if there is no critical information missing.
\\
For steps involving numerical calculations, carefully verify the accuracy of the calculations to ensure all results are correct.
\\
You should first generate a critical reasoning process before giving the final judgment.
\\
\\

\#\#Format for Evaluation\#\#
\\
For each specified step in the solution path, perform your evaluation by following the below format:
\\
**Critique of Step <current\_step>**: First generate a detailed reasoning thought to evaluate the step. 
\\
**Judgement**: Based on the above critique, give your final judgement in the form of ``\#\#\#\# The correctness of Step <current\_step> is: \verb|\|boxed\{1|-1\}'', where 1 represents correct and -1 represents incorrect. The judgement result should be either 1 or -1.
\\
\\
<Problem>
\{problem\}
</Problem>
\\
\\
<Solution Path>
\{solution\}
</Solution Path>
\\
\\
Now, please critique Step \{step\_index\} in the above solution path.
\end{prompt}
\begin{prompt}{Prompt Template for In-Depth Critique Generation}
\label{prompt: in-depth prompt generation}
You are a math expert and are tasked with evaluating the critique for a specific step in a solution to a mathematical problem.
\\
You will be given the problem, the solution path, and the critique for a specified step in the solution path.
\\
You need to critique the critique for the specified step and provide your judgement on whether the critique is correct or incorrect, and then determine the final correctness of the specified step.
\\
You need to think about how you would approach evaluating the step if you were asked to do so, without referring to the original critique.
\\
You can either re-evaluate the specified step using different valid approaches or from different perspectives than the original critique to see if different methods can reach the same conclusion; or alternatively, you can critique the original critique itself to check if it is correct and whether it is fair and reasonable.
\\
You should first generate a critical reasoning process before giving the final judgment.
\\
\\
\#\#Format for Evaluation\#\#
Perform your evaluation to the critique by following the below format:
\\
**Critique of the critique of Step <current\_step>**: First generate a detailed critique either by re-evaluating the specified step with different ways or by directly evaluating the original critique of the step.
\\
**Judgement**: Based on the results of original critique and critique's critique, give your final judgement on the correctness of the specified step in the form of ``\#\#\#\# The correctness of Step <current\_step> is: \verb|\|boxed\{1|-1\}'', where 1 represents correct and -1 represents incorrect. The judgment result should be either 1 or -1.
\\
\\
<Problem>
\{problem\}
</Problem>
\\
\\
<Solution Path>
\{solution\}
</Solution Path>
\\
\\
<Original Critique>
\{original\_critique\}
</Original Critique>
\\
\\
Now, please critique the original critique of the Step \{step\_index\} and give your final judgement on the correctness of Step \{step\_index\}.
\end{prompt}
\begin{prompt}{Prompt Template for Final Critique Synthesis}
\label{prompt: prompt for critique merging}
You are a math expert and a good math critic.
\\
You will be provided with an initial critique and a critique of the initial critique.
\\
Your task is to merge the two critiques into a single, deliberate critique.
\\
You should merge the two critiques as if they were generated in one go, as if the model first generated a critique and then wanted to further verify that step or the critique itself.
\\
You should make the merged critique smooth by adding some transitional, pausing, reflective, thinking words or sentences.
Do not use terms like ``the original critique'' as the merged critique should be considered as generated in one go.
\\
\\
Here are two examples that can serve as references for the tone and format of the merged deliberate critique:
\\
\\
<Merged Deliberate Critique Example 1>\{example1\}</Merged Deliberate Critique Example 1>
\\
\\
<Merged Deliberate Critique Example 2>\{example2\}</Merged Deliberate Critique Example 2>
\\
\\
Please follow the above examples to generate the merged deliberate critique for the below sample:
\\
\\
<Original Critique>\{original\_critique\}</Original Critique> 
\\
\\
<Critique of the Original Critique>\{critique\_of\_original\_critique\}</Critique of the Original Critique>
\end{prompt}

\section{Detailed Experimental Settings}
\subsection{Hyper-Parameters in The Data Generation Stage}
\label{appendix: hyper-parameters in critique generation}
In the initial critique generation stage, for each reasoning step in the given solution, we prompt Qwen2.5-72B-Instruct once with $\texttt{temperature}=0.7$ and $\texttt{top\_p}=0.9$. In the in-depth critique generation stage, based on the problem, solution and the initial critique, we prompt Qwen2.5-72B-Instruct 16 times with $\texttt{temperature}=1.0$ and $\texttt{top\_p}=0.9$. We then randomly select one critique from those whose judgment results $j_{i}^{deep}$ are consistent with the ground truth labels, and use it as the in-depth critique for the corresponding step. Finally, during critique merging, the sampling parameters for Qwen2.5-72B-Instruct are $\texttt{temperature}=0.7$ and $\texttt{top\_p}=0.9$.

\subsection{Overview of Evaluation Benchmarks}
\label{appendix: introduction to benchmarks}
For MR-GSM8K~\citep{Mr-gsm8k}, we only choose the subset in which the questions are from the original GSM8K~\citep{gsm8k} dataset, containing 693 correct solutions and 725 incorrect solutions. For PRM800K~\citep{verify_step_by_step}, we adopt the Phase-2 test set that contains 586 correct solutions and 2078 incorrect solutions. The third evaluation benchmark is ProcessBench~\citep{process-and_outcome-based_feedback}, which includes four subsets in which the questions originates from different sources: GSM8K, MATH~\citep{math}, OlympiadBench~\citep{olympiadbench} and Omni-Math~\citep{omni-math}. The total number of solutions for these four subsets are 400, 1000, 1000, and 1000, respectively.

\subsection{Detailed Training Settings}
\label{appendix: training setting}
The complete training hyper-parameters in SFT and RL are put in Table~\ref{tab: sft hyper-parameters} and Table~\ref{tab: rl hyper-parameters} respectively. 
In RL, an accuracy reward of 1.0 is given if the final judgment is correct; otherwise, it is 0.0.

\begin{figure}[t]  
\begin{center}
\begin{minipage}[t]{0.48\linewidth}
\caption{Training hyper-parameters in SFT.}
\label{tab: sft hyper-parameters}
\centering
\begin{tabular}{ll}
\toprule
Hyper-parameter & Value \\
\midrule
LR & $1\times 10^{-5}$  \\
LR Scheduler & cosine \\
Batch Size & 64 \\
Epochs & 3 \\
Maximum Sequence Length & 16384 \\
Warmup Ratio & 0.1 \\
\bottomrule
\end{tabular}
\end{minipage}  
\hfil
\begin{minipage}[t]{0.48\linewidth}
\caption{Training hyper-parameters in RL.}
\label{tab: rl hyper-parameters}
\centering
\begin{tabular}{ll}
\toprule
Hyper-parameter & Value \\
\midrule
Train Batch Size & 128  \\
Micro Batch Size & 128 \\
Rollout $n$ & 8 \\
Maximum Prompt Length & 2048 \\
Maximum Response Length & 8192 \\
Temperature & 1.0 \\
Top $p$ & 0.9 \\
LR & $1\times 10^{-6}$ \\
Epochs & 2 \\
KL Coefficient & 0.0 \\
\bottomrule
\end{tabular}
\end{minipage} 
\end{center}
\end{figure}

\subsection{Detailed Evaluation Settings}
\label{appendix: evaluation settings}
In the main evaluations (Table~\ref{tab: main results}), we use consistent sampling settings across all critique models, with \texttt{temperature} set to 0.6, \texttt{top\_p} to 0.9, and \texttt{max\_generation\_length} to 32K during inference. We only sample once for each task input in this setting. We then calculate the judgment accuracy on the step index of first erroneous step in incorrect solutions and the judgment accuracy on correct solutions. The F1 score is then calculated as the harmonic mean of the above two numbers.

In the experiments of majority voting (Section~\ref{subsec: results of majority voting}), for each selected critique model, we sample 8 responses with \texttt{temperature} set to 1.0 and \texttt{top\_p} to 0.9.

The evaluation prompt is mainly based on that used in~\citep{Processbench}, and is put in below. We do not directly use the prompt used in~\citep{Processbench}, because we find that it will make the model tend to directly generate the judgment results without the reasoning process.

\begin{prompt}{Evaluation Prompt}
\label{prompt: evaluation prompt}
The following is a math problem and a solution.
\\
The solution is presented as a step-by-step chain of thought.
\\
Each step is separated with the index indicator ``Step i:'', with i indexed starting from 1.
\\
\\
<Problem>
\\
\{problem\}
\\
</Problem>
\\
\\
<Solution>
\\
\{tagged\_response\}
\\
</Solution>
\\
\\
Your task is to evaluate the solution and identify the earliest step that contains an error, or confirm that all steps are correct.
\\
Please first review and generate a critique for each step.
\\
After reviewing each step, once you identify an error in the step, stop the critique and return the index of that step as this is the step where the earliest error occurs. Otherwise, continue reviewing the subsequent steps. If all steps are correct, return the index of -1 (which typically denotes ``not found'').
\\
Finally, put your final answer (i.e., the step index) in \verb|\|boxed\{\}.
\end{prompt}

\begin{figure}[h] 
    \centering
    \includegraphics[width=\textwidth]{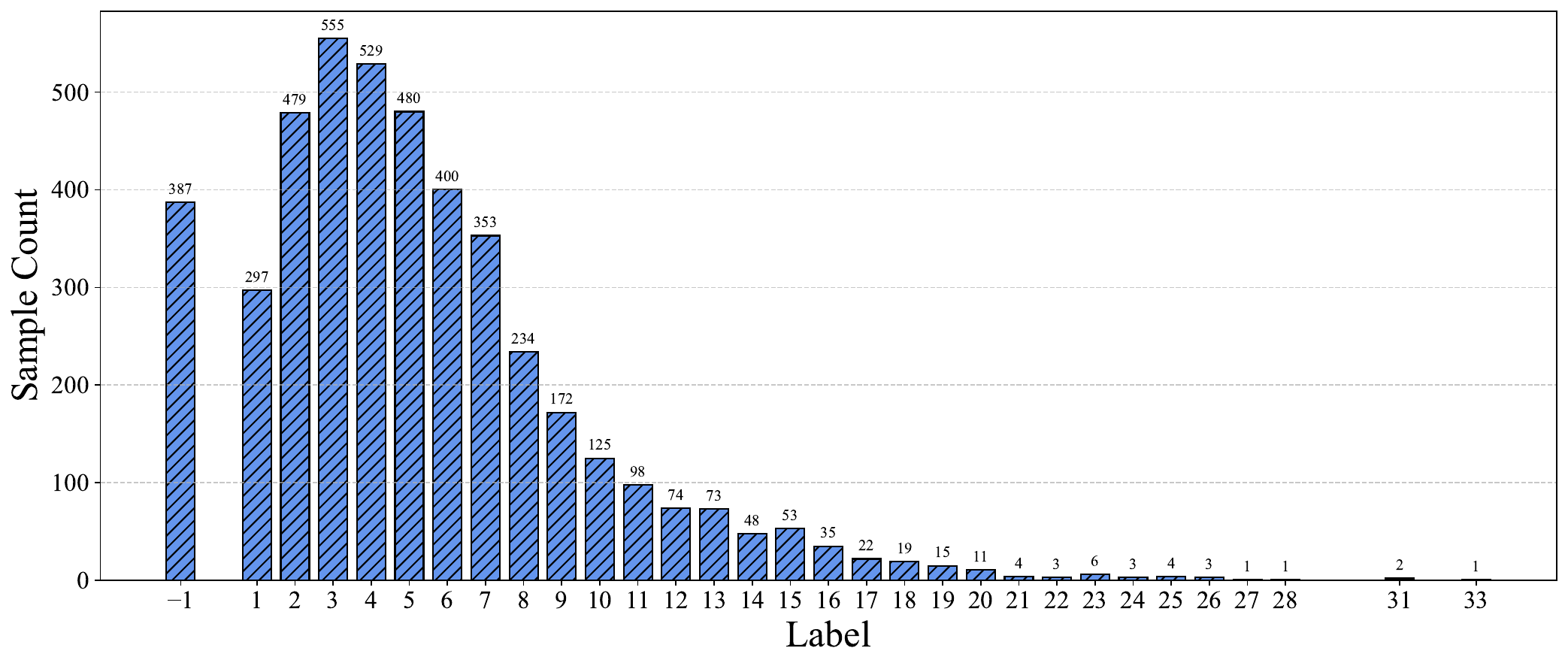}
    \caption{Solution-level label distribution of seed critique data in SFT stage.
    }
    \label{fig: label distribution in sft data}
\end{figure}

\begin{figure}[h]  
\begin{center}
\begin{minipage}[t]{0.48\linewidth}
\centerline{\includegraphics[width=1\linewidth]{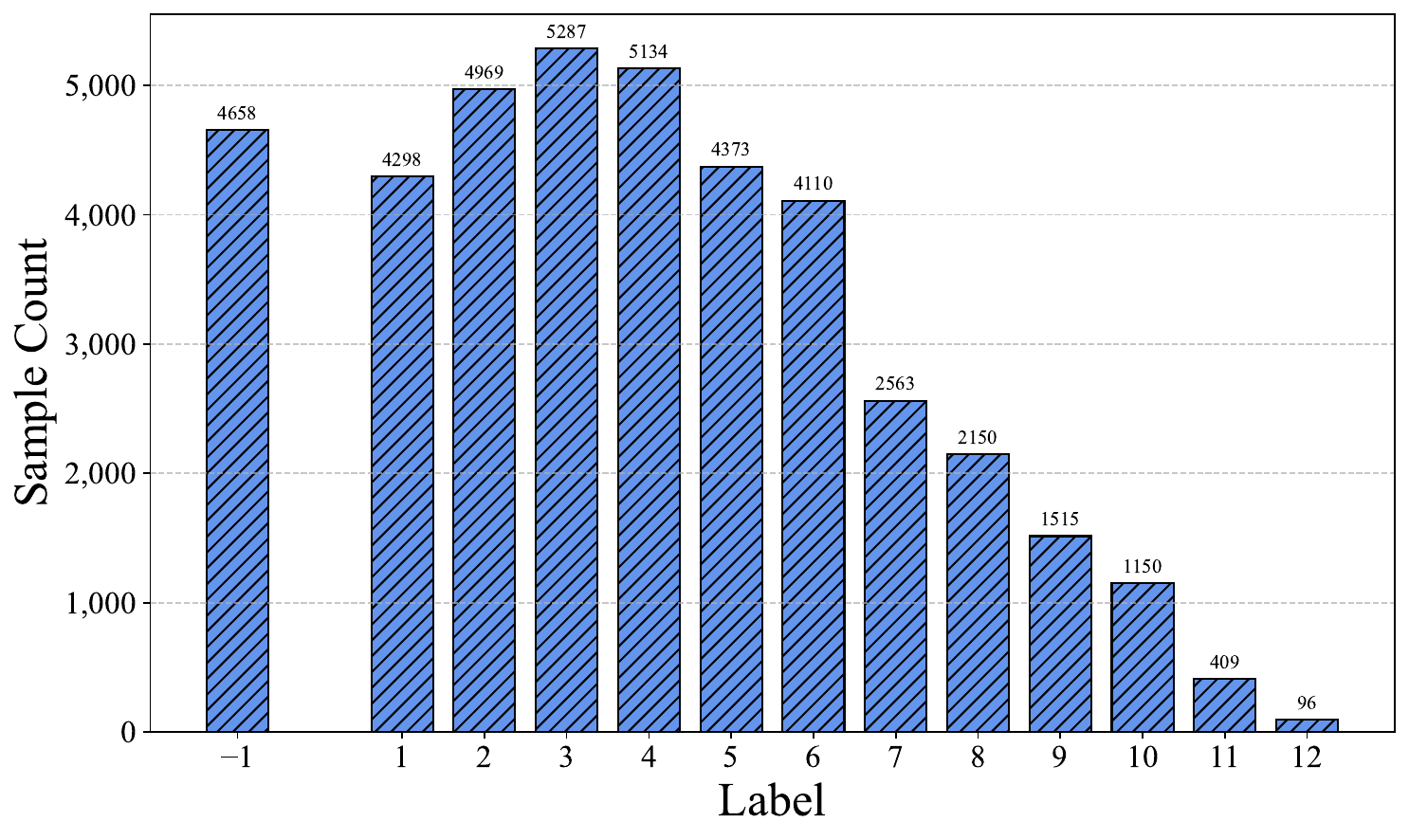}}
\caption{Solution-level label distribution of RL data based on PRM800K.}
\label{fig: label distribution in prm800k data}
\end{minipage}  
\hfill
\begin{minipage}[t]{0.48\linewidth}
\centerline{\includegraphics[width=1\linewidth]{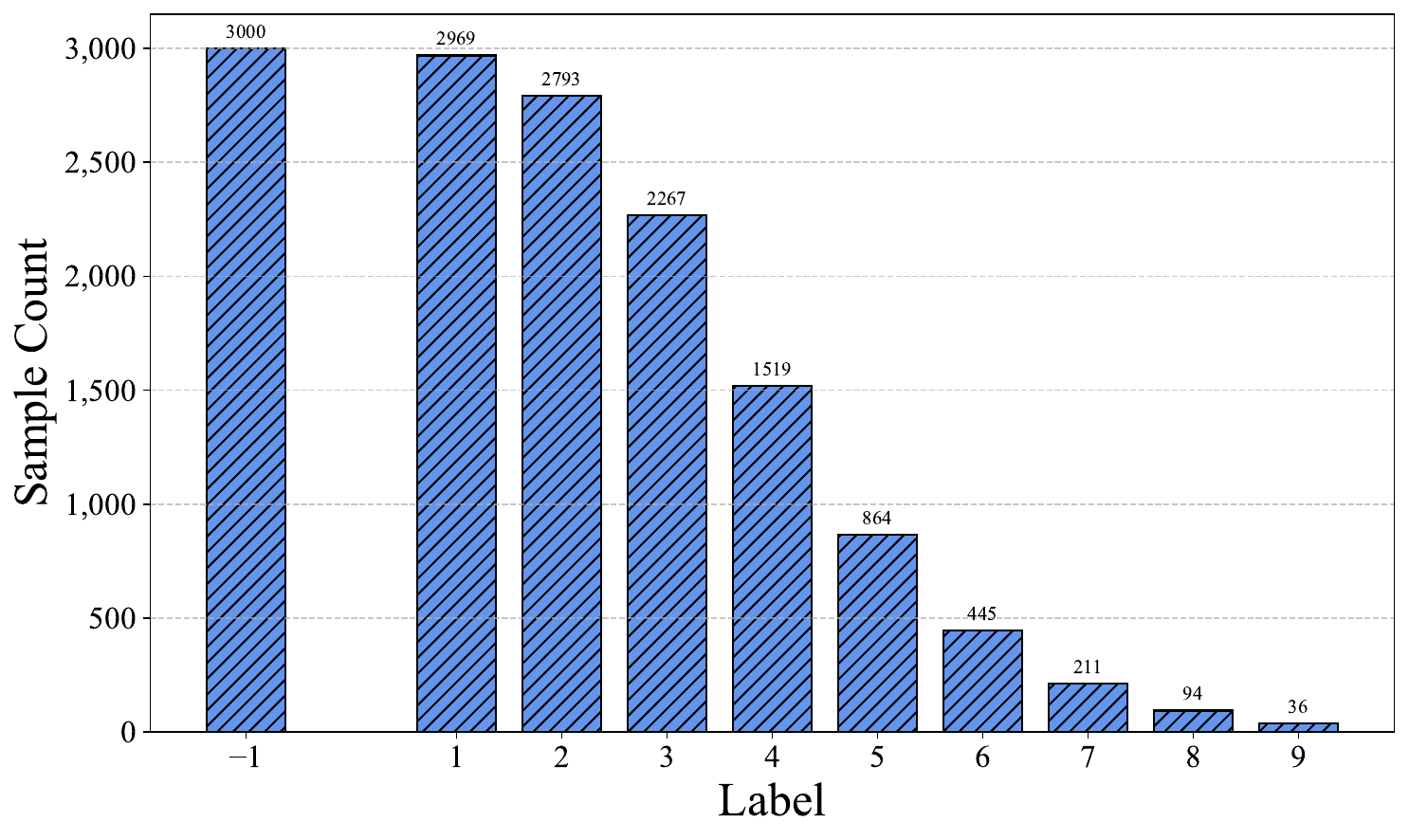}}
\caption{Solution-level label distribution of RL data based on NuminaMath-CoT.}
\label{fig: label distribution in numina data}
\end{minipage}  
\end{center}
\end{figure}

\section{Label Distributions of SFT and RL Data}
We show the solution-level label distribution (i.e., he distribution of the step index of the first erroneous step) of our curated SFT dataset in Figure~\ref{fig: label distribution in sft data}. The label distributions of RL data based on PRM800K and NuminaMath-CoT are show in Figure~\ref{fig: label distribution in prm800k data} and Figure~\ref{fig: label distribution in numina data} respectively. We note that due to limited computational resources, we set the \texttt{max\_response\_length} to 8192 during the RL phase. As a result, \textbf{we apply filtering and constraints to the RL data, retaining only task inputs whose solutions contain a number of reasoning steps within a certain range}. This avoids the situation where a large number of rollouts would be truncated due to excessive lengths, which could degrade training performance. However, we believe that with sufficient computational resources, incorporating inputs with more steps and increasing the \texttt{max\_response\_length} during RL can further improve RL performance.

\section{Prompt Templates in Verified Majority Voting and Critique-Based Refinement Experiments}
\label{appendix: refinement experiments settings}

The prompt templates to create step-by-step solutions and critique-based refinements for generators are put below separately. In the experiments of critique-based refinements, we further add instructions to instruct DeepSeek-R1-Distill-Qwen-7B not to produce the final answer and to stop critiquing after identifying the first incorrect step. However, even with this constraint, we still find that in a certain number of cases, DeepSeek-R1-Distill-Qwen-7B does not follow the instruction and continues critiquing until the end.

\begin{prompt}{Prompt Template for Generating Step-by-Step Solutions}
\label{prompt: prompt for step-by-step solutions}
System: You are a helpful assistant. Please reason step by step, and put your final answer within \verb|\|boxed\{\}.
\\
User: \{problem\}
\\
Assistant: Step 1:
\end{prompt}

\begin{prompt}{Prompt Template for Critique-Based Refinement}
\label{prompt: refinement prompt}
System: You are a helpful assistant. Please reason step by step, and put your final answer within \verb|\|boxed\{\}.
\\
User: \{problem\}
\\
Assistant: \{initial\_solution\}
\\
User: There might be some problems in your solution. Here is the critique of the above solution:
\\
\{critique\}
\\
Please carefully refine the solution based on the critique.
\\
Assistant:
\end{prompt}

\section{Case Study}
\begin{figure*}[t!] 
    \centering
    \includegraphics[width=\textwidth]{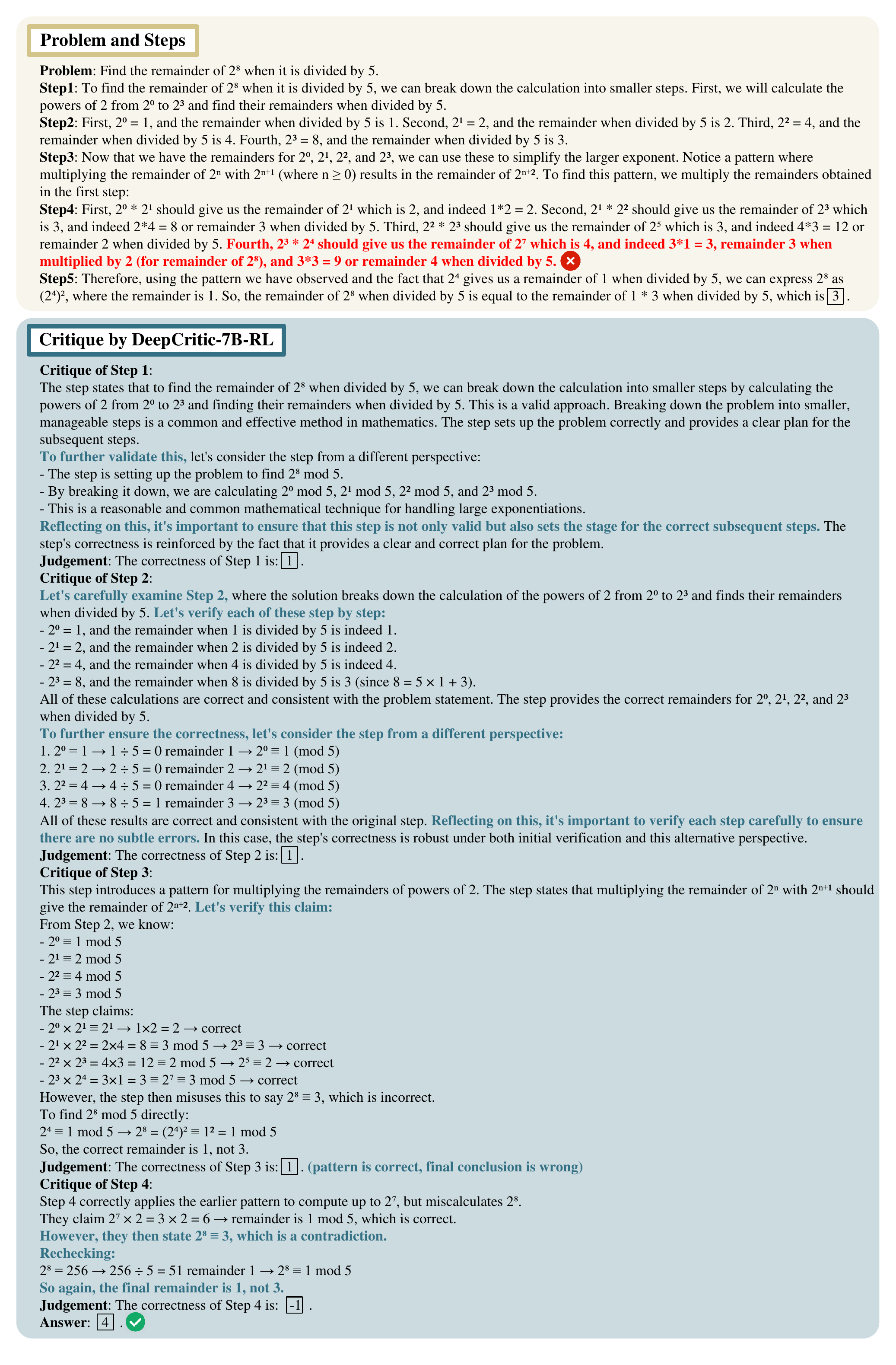}
    \caption{A complete case study of DeepCritic-7B-RL-PRM800K.
    }
    \label{fig: case}
\end{figure*}

We present a complete case study of our DeepCritic-7B-RL-PRM800K model in Figure~\ref{fig: case}.

\label{appendix: case study}

\section{Detailed Results on All Benchmarks}
\label{appendix: detailed results on all benchmarks}
We present the detailed results—including the separate judgment accuracy on both correct and incorrect solutions—on each benchmark in Table~\ref{tab: detailed results on mr-gsm8k and prm800k},~\ref{tab: detailed results on gsm8k and math}, and~\ref{tab: detailed results on olympiad bench and omnimath}.

\begin{table*}[t]
\caption{Detailed results of various models on MR-GSM8K and PRM800K. ``erroneous'' represents the judgment accuracy on incorrect solutions, ``correct'' represents the judgment accuracy on correct solutions, ``F1'' represents the harmonic mean.}
\label{tab: detailed results on mr-gsm8k and prm800k}
\centering
\setlength{\tabcolsep}{4pt}
\begin{tabular}{lcccccc}
\toprule
\multirow{2.5}{*}{\begin{tabular}[c]{@{}l@{}}Model \end{tabular}} & \multicolumn{3}{c}{MR-GSM8K} & \multicolumn{3}{c}{PRM800K}  \\
\cmidrule(lr){2-4}
\cmidrule(lr){5-7}
&   erroneous & correct &  F1  &  erroneous & correct &  F1   \\
\midrule
\multicolumn{7}{l}{\emph{\quad \textbf{Process Reward Models} (PRMs)}} \\
Math-Shepherd-PRM-7B & 48.8 & 84.1 & 61.8 & 14.0 & 47.8 & 21.7\\
RLHFlow-PRM-8B-Mistral & 50.8 & 96.7 & 66.6 & 15.6 & 65.0 & 25.2  \\
RLHFlow-PRM-8B-DeepSeek & 29.0 & 99.0 & 44.8 & 10.3 & 89.6 & 18.5 \\
Qwen2.5-Math-7B-PRM800K &  58.8 & 89.2 & 70.8 & 43.4 & 77.3 & 55.6 \\
\midrule
\multicolumn{7}{l}{\emph{\quad Large Language Models, served as \textbf{Critique Models}}} \\
LLaMA3.1-8B-Instruct & 35.0 & 28.7 & 31.6 & 17.2 & 15.0 & 16.0 \\
Qwen2.5-7B-Instruct & 33.4 &  86.0 & 48.1 & 15.2 & 81.7 & 25.6\\
Qwen2.5-7B-Instruct (Maj@8) & 34.9 & 89.8 & 50.3 & 17.1 & 82.8 & 28.3 \\
Qwen2.5-Math-7B-Instruct & 23.0 & 78.5 & 35.6 & 10.9 & 88.9 & 19.4 \\
DeepSeek-R1-Distill-Llama-8B & 59.0 & 84.3 & 69.4 & 44.3 & 75.3 & 55.7\\
DeepSeek-R1-Distill-Qwen-7B & 67.3 & 92.5 & 77.9 &  42.3 & 89.2 & 57.4\\
DeepSeek-R1-Distill-Qwen-7B (Maj@8) & 69.4 & 95.2 & 80.3 & 49.6 & 92.5 & 64.5 \\
LLaMA3.1-70B-Instruct &65.0 & 81.7 & 72.4 & 21.9 &  76.8 & 34.1 \\
Qwen2.5-72B-Instruct & 59.7 & 92.6 & 72.6 & 31.0 & 84.3 & 45.3 \\
Qwen2.5-Math-72B-Instruct & 61.1 & 92.6 & 73.6 & 26.6 & 89.8 & 41.0 \\
GPT-4o & 61.8 & 80.0 & 69.7 & 31.5 & 84.3 & 45.9\\
\midrule
\multicolumn{7}{l}{\emph{\quad \textbf{Our Critique Models}}} \\
DeepCritic-7B-SFT & 55.3 & 85.1 &  67.1 & 35.0 & 76.3 & 48.0 \\
DeepCritic-7B-RL-Numina & 65.4 & 94.2 & 77.2 & 45.4 & 72.5 & 55.9 \\
DeepCritic-7B-RL-PRM800K &  66.2 & 92.9 & 77.3 & 49.6 & 76.3 & 60.1 \\
DeepCritic-7B-RL-PRM800K (Maj@8) & 68.0 & 93.4 & 78.7 & 53.7 & 81.1 & 64.6\\
\bottomrule
\end{tabular}
\end{table*}

\begin{table*}[t]
\caption{Detailed results of various models on the GSM8K and MATH testing sets of ProcessBench. ``erroneous'' represents the judgment accuracy on incorrect solutions, ``correct'' represents the judgment accuracy on correct solutions, ``F1'' represents the harmonic mean.}
\label{tab: detailed results on gsm8k and math}
\setlength{\tabcolsep}{4pt}
\centering
\begin{tabular}{lcccccc}
\toprule
\multirow{2.5}{*}{\begin{tabular}[c]{@{}l@{}}Model \end{tabular}} & \multicolumn{3}{c}{GSM8K} & \multicolumn{3}{c}{MATH}  \\
\cmidrule(lr){2-4}
\cmidrule(lr){5-7}
&   erroneous & correct & F1  &  erroneous & correct &  F1   \\
\midrule
\multicolumn{7}{l}{\emph{\quad \textbf{Process Reward Models} (PRMs)}} \\
Math-Shepherd-PRM-7B & 32.4 & 94.3 & 48.2 & 16.2 & 83.7 & 27.1\\
RLHFlow-PRM-8B-Mistral & 34.3 & 98.4 & 50.9 & 20.0 & 79.3 & 32.0\\
RLHFlow-PRM-8B-DeepSeek & 19.3 & 99.0 & 32.3 & 21.7 & 80.8 & 34.2\\
Qwen2.5-Math-7B-PRM800K & 56.5 & 93.8 & 70.5 & 50.8 & 88.9 & 64.7 \\
\midrule
\multicolumn{7}{l}{\emph{\quad Large Language Models, served as \textbf{Critique Models}}} \\
LLaMA3.1-8B-Instruct & 36.7 & 17.6 & 23.8 & 23.7 & 15.8 & 18.9\\
Qwen2.5-7B-Instruct & 29.0 & 82.4 & 42.9 & 23.2 & 86.0 & 36.6 \\
Qwen2.5-7B-Instruct (Maj@8) & 33.3 & 91.2 & 48.8 & 24.6 & 89.2 & 38.5 \\
Qwen2.5-Math-7B-Instruct & 13.0 & 99.5 & 23.1 & 12.5 & 94.1 & 22.0\\
DeepSeek-R1-Distill-Llama-8B & 55.1 & 79.3 & 65.0 & 58.4 & 67.7 & 62.7 \\
DeepSeek-R1-Distill-Qwen-7B & 57.0 & 97.4 & 71.9 & 57.7 & 88.4  & 69.9 \\
DeepSeek-R1-Distill-Qwen-7B (Maj@8) & 65.7 & 99.5 & 79.1 & 65.7 & 92.9 & 76.9 \\
LLaMA3.1-70B-Instruct & 61.4 & 88.6 & 72.5 & 33.7 & 81.0 & 47.6\\
Qwen2.5-72B-Instruct &  57.5 & 96.9 & 72.2 & 36.2 & 94.8 & 52.4\\
Qwen2.5-Math-72B-Instruct & 53.1 &  96.9 & 68.6 & 32.5 & 95.3 & 48.5 \\
GPT-4o & 58.0 & 95.3 & 72.1 & 41.2 & 94.3 & 57.3\\
\midrule
\multicolumn{7}{l}{\emph{\quad \textbf{Our Critique Models}}} \\
DeepCritic-7B-SFT & 43.5 & 92.7 & 59.2 &  47.0 & 87.7 & 61.2\\
DeepCritic-7B-RL-Numina & 56.5 & 94.3 & 70.7 & 57.4 & 77.3 & 65.9 \\
DeepCritic-7B-RL-PRM800K & 60.9 & 94.3 & 74.0 & 64.3 & 84.2 & 72.9 \\
DeepCritic-7B-RL-PRM800K (Maj@8) & 64.7 & 97.4 &  77.8 & 67.9 &85.5 & 75.6 \\
\bottomrule
\end{tabular}
\end{table*}

\begin{table*}[t]
\caption{Detailed results of various models on the OlympiadBench and Omni-Math testing sets of ProcessBench. ``erroneous'' represents the judgment accuracy on incorrect solutions, ``correct'' represents the judgment accuracy on correct solutions, ``F1'' represents the harmonic mean.}
\label{tab: detailed results on olympiad bench and omnimath}
\setlength{\tabcolsep}{4pt}
\centering
\begin{tabular}{lcccccc}
\toprule
\multirow{2.5}{*}{\begin{tabular}[c]{@{}l@{}}Model \end{tabular}} & \multicolumn{3}{c}{OlympiadBench} & \multicolumn{3}{c}{Omni-Math}  \\
\cmidrule(lr){2-4}
\cmidrule(lr){5-7}
&   erroneous & correct &  F1  &  erroneous & correct &  F1   \\
\midrule
\multicolumn{7}{l}{\emph{\quad \textbf{Process Reward Models} (PRMs)}} \\
Math-Shepherd-PRM-7B & 11.8 & 78.8 & 20.5 & \phantom{0}9.1 & 79.3 & 16.3\\
RLHFlow-PRM-8B-Mistral & \phantom{0}8.0 & 48.4 & 13.8 & \phantom{0}9.4 & 49.4 & 15.7\\
RLHFlow-PRM-8B-DeepSeek & \phantom{0}9.4 & 55.5 & 16.0 & 11.1 & 52.7 & 18.3\\
Qwen2.5-Math-7B-PRM800K & 35.4 & 85.0 & 50.0 & 28.7 & 83.4 & 42.7 \\
\midrule
\multicolumn{7}{l}{\emph{\quad Large Language Models, served as \textbf{Critique Models}}} \\
LLaMA3.1-8B-Instruct & 20.4 & 16.5 & 18.3 & 20.9 & 14.5 & 17.2\\
Qwen2.5-7B-Instruct & 15.0 & 86.4 & 25.5 & 15.4 & 80.9 & 25.9 \\
Qwen2.5-7B-Instruct (Maj@8) & 16.0 & 90.9 & 27.3 & 18.1 & 86.3 & 29.9 \\
Qwen2.5-Math-7B-Instruct & \phantom{0}4.8 & 85.8 & \phantom{0}9.2 & \phantom{0}5.5 & 88.0 & 10.4\\
DeepSeek-R1-Distill-Llama-8B & 50.4 & 69.3 & 58.4 & 40.6 & 71.4 &   51.7\\
DeepSeek-R1-Distill-Qwen-7B & 43.6 & 79.9 & 56.4 & 33.9 & 75.5 & 46.8 \\
DeepSeek-R1-Distill-Qwen-7B (Maj@8) & 50.5 & 87.3 & 64.0 & 40.8 & 81.3 & 54.4\\
LLaMA3.1-70B-Instruct & 29.5 & 67.3 & 41.0 & 25.6 & 65.6 & 36.8\\
Qwen2.5-72B-Instruct & 26.9 & 94.4 & 41.9 & 28.1 & 92.5 & 43.1 \\
Qwen2.5-Math-72B-Instruct & 16.8 & 96.8 & 28.6 & 15.9 & 95.0 & 27.3 \\
GPT-4o & 35.3 & 88.4 & 50.5 & 39.2 & 83.8 & 53.4\\
\midrule
\multicolumn{7}{l}{\emph{\quad \textbf{Our Critique Models}}} \\
DeepCritic-7B-SFT & 32.8 & 76.7 & 46.0 & 30.2 & 75.1 & 43.0 \\
DeepCritic-7B-RL-Numina & 49.8 & 68.4 & 57.6 & 45.8 & 64.3 & 53.5 \\
DeepCritic-7B-RL-PRM800K & 52.3 & 72.9 &  60.9  & 48.0 & 71.0 & 57.2 \\
DeepCritic-7B-RL-PRM800K (Maj@8) & 59.8 & 75.2 & 66.6 & 52.3 & 68.9 & 59.5 \\
\bottomrule
\end{tabular}
\end{table*}

\end{document}